\documentclass[letterpaper, 10 pt, conference]{ieeeconf}  

\IEEEoverridecommandlockouts                              

\usepackage[nolist]{acronym} 
\usepackage{amsmath,amsfonts,amssymb,siunitx,booktabs,cite} 
\let\labelindent\relax
\usepackage[inline]{enumitem}
\usepackage{subcaption}
\usepackage{caption}
\usepackage{graphicx}
\usepackage[usenames,dvipsnames]{xcolor}
\usepackage{tikz,standalone,pgfplots,tikzscale}
\usepackage[hidelinks]{hyperref} 
\pgfplotsset{compat=1.8}
\usetikzlibrary{positioning}
\usepackage{wrapfig}
\usepackage{textcomp}
\usepackage{overpic}
\usepackage[linecolor=blue!50,backgroundcolor=blue!10,textsize=tiny,disable]{todonotes} 
\usepackage{pifont}
\usepackage{siunitx}
\usepackage{flushend}
\newcommand{\cmark}{\ding{51}}%
\newcommand{\xmark}{\ding{55}}%
\usepackage[hypcap=false]{caption}
\usepackage{float}
\newacro{dnn}[DNN]{Deep Neural Network}
\newacro{fcn}[FCN]{Fully Convolutional Network}
\newacro{pc}[PC]{Point Cloud}
\newacro{sdf}[SDF]{Signed Distance Function}
\newacro{cnn}[CNN]{Convolutional Neural Network}
\newacro{gnn}[GNN]{Graph Neural Network}
\newacro{dl}[DL]{Deep Learning}
\newacro{ml}[ML]{Machine Learning}
\newacro{mc}[MC]{Monte Carlo}
\newacro{mlp}[MLP]{Multi-Layer Perceptron}
\newacro{dof}[DoF]{Degrees of Freedom}
\newacro{vae}[VAE]{Variational Autoencoder}
\newacro{cvae}[CVAE]{Conditional Variational Autoencoder}
\newacro{methodname}[VCGS]{Variational Constrained Grasp Sampler}
\newacro{fps}[FPS]{Farthest Point Sampling}
\newacro{tai}[TaI]{Target as Input}
\newacro{elbo}[ELBO]{Evidence Lower Bound}

\newcommand{\equationref}[1]{\hyperref[#1]{Eq.~\ref*{#1}}}
\newcommand{\figref}[1]{\hyperref[#1]{Fig.~\ref*{#1}}}
\newcommand{\tabref}[1]{\hyperref[#1]{Table~\ref*{#1}}}
\newcommand{\secref}[1]{\hyperref[#1]{Section~\ref*{#1}}}
\newcommand{\algoref}[1]{\hyperref[#1]{Algorithm~\ref*{#1}}}

\newcommand{\norm}[1]{\left\lVert#1\right\rVert}

\newcommand{\ra}[1]{\renewcommand{\arraystretch}{#1}}
\newcommand{\tbs}[1]{\renewcommand{\tabcolsep}{#1pt}}

\newcommand{\matr}[1]{\mathbf{#1}}

\newcommand*{\prob}{\mathsf{P}}

\newcommand{\etal}[1]{#1 et al.}

\def\methodname{VCGS}
\def\datasetname{CONG}
\def\franka{Franka Emika Panda}
\def\kinect{Kinect 2.0}
\def\graspnet{GraspNet}
\def\graspnetta{GraspNet \ac{tai}}
\def\isaacgym{Isaac Gym}

\def\sota{state-of-the-art}
\def\ie{, \textit{i.e.}, }

\def\pc{point cloud}
\def\pcs{point clouds}

\title{\LARGE \bf
Constrained Generative Sampling of 6-DoF Grasps
}

\author{Jens~Lundell$^{\dagger}$, Francesco~Verdoja$^{*}$, Tran~Nguyen~Le$^{*}$, Arsalan~Mousavian$^{\ddagger}$, Dieter~Fox$^{\ddagger,\mathsection}$ and Ville~Kyrki$^{*}$
\thanks{$^{*}$ Intelligent Robotics Group, Department of Electrical Engineering  and Automation, School of Electrical Engineering, Aalto University. Finland.}%
\thanks{$^{\dagger}$ KTH Royal Institute of Technology, Sweden {\tt\small jelundel@kth.se}.}%
\thanks{$^{\ddagger}$ NVIDIA Corporation, USA}
\thanks{$^{\mathsection}$ Paul G. Allen School of Computer Science \& Engineering, University of Washington, Seattle, USA}%
}

\makeatletter
\let\@oldmaketitle\@maketitle
\renewcommand{\@maketitle}{\@oldmaketitle
  \setcounter{figure}{0}
  \vspace{1em}
  
    \begin{tikzpicture}
        \node (pull_fig_0) [label=above:{\large\textbf{Setup}}] {\includegraphics[width=8cm]{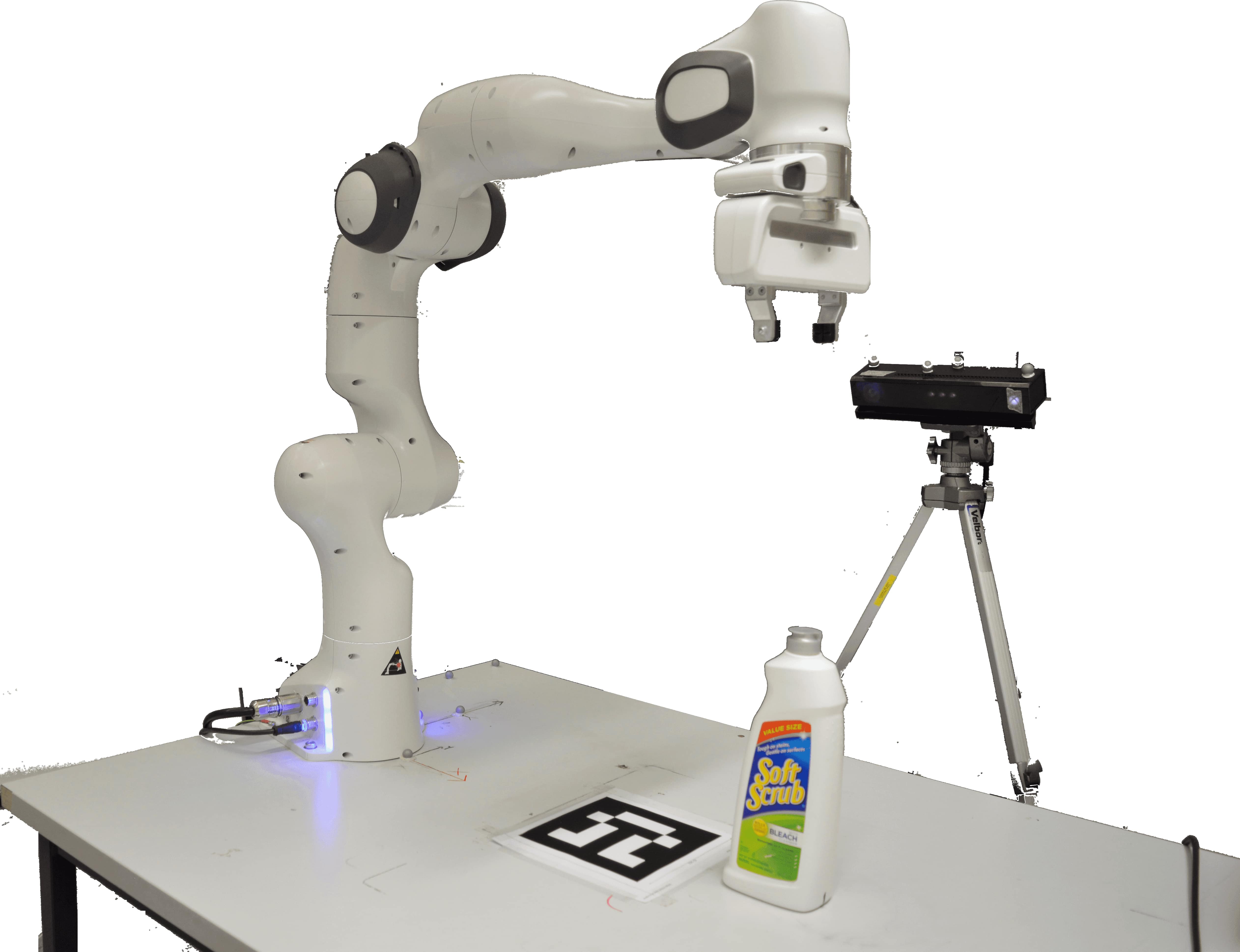}};
        \node (pull_fig_2) [right=of pull_fig_0, label={[xshift=-0.6cm, yshift=0.45cm]above:{\large\textbf{Segmentation}}}] {\scalebox{-1}[1]{\includegraphics[width=2cm]{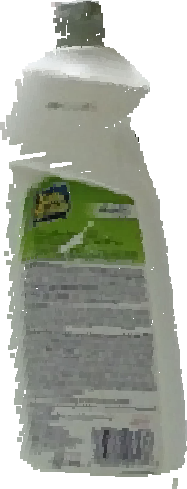}}};
        \node (pull_fig_3) [right=of pull_fig_2, label={[xshift=-0.5cm, yshift=0.45cm]above:{\large\textbf{Down-sampling}}}] {\scalebox{-1}[1]{\includegraphics[width=2cm]{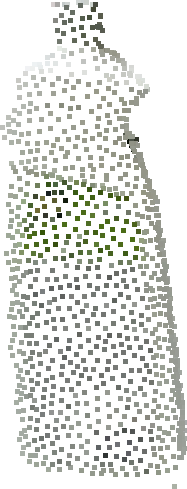}}};
        \node (pull_fig_4) [right=of pull_fig_3, label={[xshift=0.3cm, yshift=0.42cm]above:{\large\textbf{Selected target area}}}] {\scalebox{-1}[1]{\includegraphics[width=2cm]{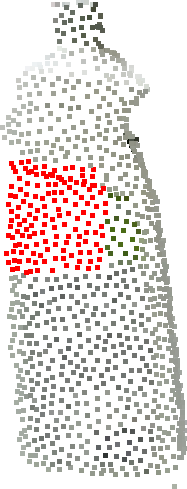}}};
        \node (pull_fig_5) [right=of pull_fig_4, label={[xshift=0.6cm, yshift=0.38cm]above:{\large\textbf{Generated grasps}}}] {\scalebox{-1}[1]{\includegraphics[width=3.8cm]{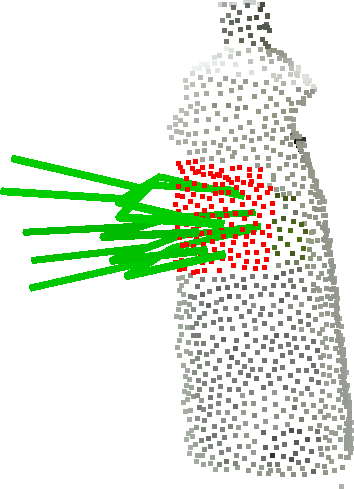}}};
        \node (pull_fig_6) [right=of pull_fig_5, , label={[yshift=0.75cm]above:{\large\textbf{Executed grasp}}}] {\includegraphics[width=8cm]{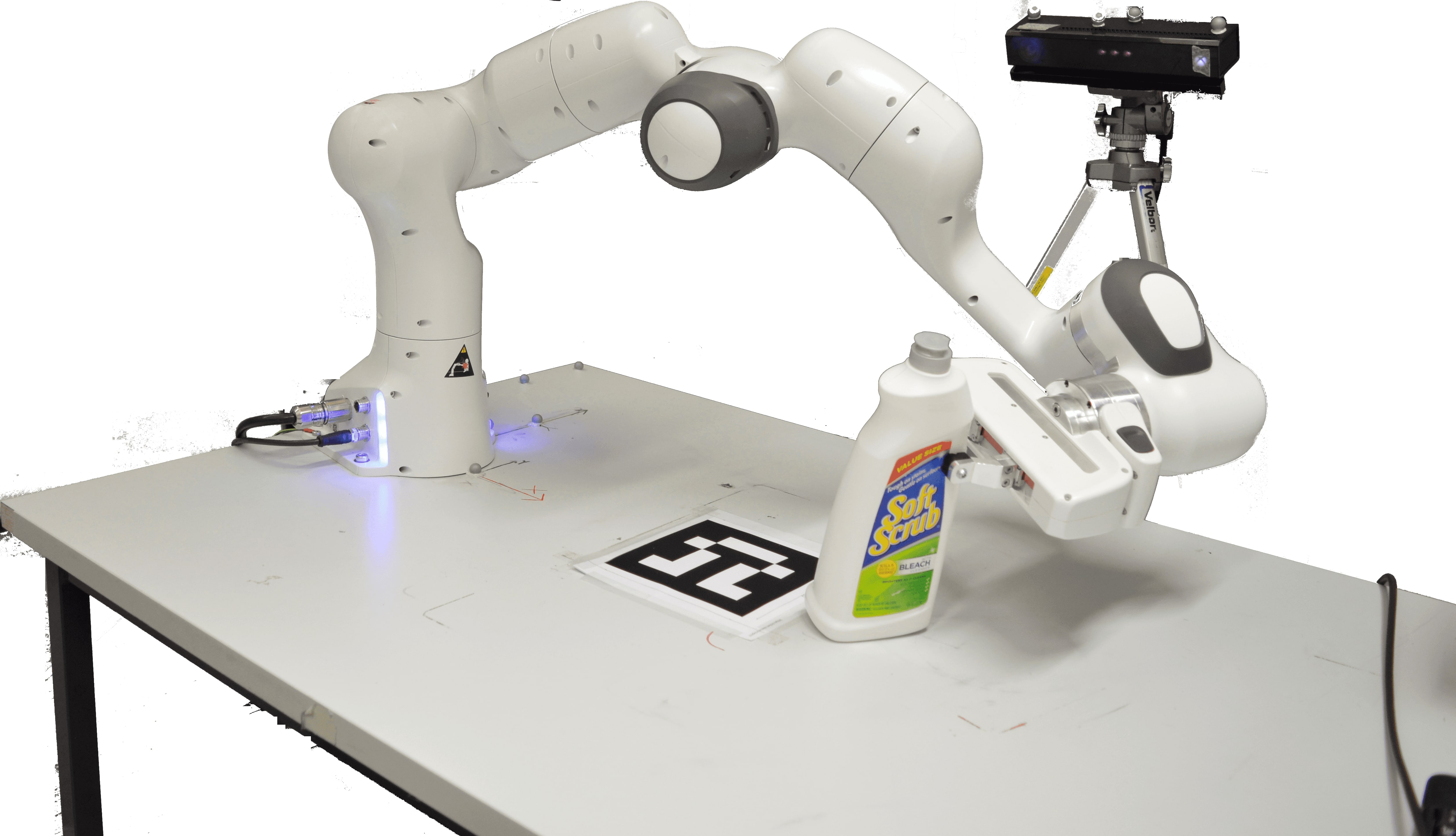}};

        \draw [line width=1mm,black,->] (pull_fig_0) to[left] (pull_fig_2);
        \draw [line width=1mm,black,->] (pull_fig_2) to[left] (pull_fig_3);
        \draw [line width=1mm,black,->] (pull_fig_3) to[left] (pull_fig_4);
        \draw [line width=1mm,black,->] (pull_fig_4) to[left] (pull_fig_5);
        \draw [line width=1mm,black,->] (pull_fig_5) to[left] (pull_fig_6);

\end{tikzpicture}

  \captionof{figure}{An example grasp generated by VCGS on the target grasping area highlighted in red. \label{fig:pull_figure} 
}}
\makeatother
\begin{document}

\maketitle

\thispagestyle{empty}
\pagestyle{empty}


\begin{abstract}

Most \sota{} data-driven grasp sampling methods propose stable and collision-free grasps uniformly on the target object. For bin-picking, executing any of those reachable grasps is sufficient. However, for completing specific tasks, such as squeezing out liquid from a bottle, we want the grasp to be on a specific part of the object's body while avoiding other locations, such as the cap. This work presents a generative grasp sampling network, \methodname{}, capable of constrained 6-\ac{dof} grasp sampling. In addition, we also curate a new dataset designed to train and evaluate methods for constrained grasping. The new dataset, called \datasetname{}, consists of over 14 million training samples of synthetically rendered point clouds and grasps at random target areas on 2889 objects. \methodname{} is benchmarked against \graspnet{}, a \sota{} unconstrained grasp sampler, in simulation and on a real robot. The results demonstrate that \methodname{} achieves a 10--15\% higher grasp success rate than the baseline while being 2--3 times as sample efficient. Supplementary material is available on \href{https://constrained-generative-sampling.github.io/exp/}{{\color{blue}our project website}}.

\end{abstract}

\section{Introduction}
\label{sec:introduction}

Most \sota{} data-driven grasp sampling methods \cite{mahlerDexNetDeepLearning2017a,morrisonClosingLoopRobotic2018b,mousavian6DOFGraspNetVariational2019c,sundermeyerContactGraspNetEfficient6DoF2021b}  focus on generating stable and collision-free grasps uniformly on the target object, which works well for completing bin-picking tasks. However, completing other tasks, such as squeezing the liquid from a bottle shown in \figref{fig:pull_figure}, often requires grasping specific target areas. A possible approach to use \sota{} grasp sampling methods to generate grasps on specific target areas is to filter out grasps outside those areas. Unfortunately, this option is extremely sample-inefficient as it requires sampling many grasps to ensure that some high-quality ones are kept. Another option is to constrain grasp sampling to specific target regions \cite{kokicLearningTaskOrientedGrasping2020a,detryTaskorientedGraspingSemantic2017,muraliSameObjectDifferent2021,liu2020cage}. In comparison to the first option, the second one has the promise of being much more sample efficient. Unfortunately, these constrained grasp sampling methods focus on generating grasps that either fulfill a specific task \cite{muraliSameObjectDifferent2021} or are located at semantically meaningful areas \cite{kokicLearningTaskOrientedGrasping2020a,detryTaskorientedGraspingSemantic2017,liu2020cage}. In this work, we do not make these assumptions and instead propose a general constrained grasp sampling method capable of focusing grasp sampling on \textit{any} target area on the object, as demonstrated in \figref{fig:pull_figure}. 

Towards learning a constrained grasp sampler, we propose the \ac{methodname}: a new generative 6-\ac{dof} constrained grasp sampling method. \methodname{} takes as input a \pc{} of the object to grasp and the target area and produces multiple 6-\ac{dof} grasps around the target area. We also curate a new dataset, \datasetname{}, to train \methodname{}. \datasetname{} consists of synthetically rendered \pcs{} of 2889 objects, and over 37 million grasps constrained to randomly sampled target areas. 

We empirically evaluated \methodname{} in terms of grasp success rates and sample efficiency by benchmarking it against the \sota{} unconstrained 6-\ac{dof} grasp sampler \graspnet{} \cite{mousavian6DOFGraspNetVariational2019c} on 126 objects in simulation and 12 objects in the real world. The experimental results demonstrate that the proposed constrained grasp sampler is 2--3 times as sample efficient as the unconstrained sampler while attaining 10--15\% higher grasp success rates.  

The main contributions of this work are:
\begin{itemize}
    \item The \acf{methodname}: a novel constrained 6-\ac{dof} generative grasp sampling deep neural network. 
    \item \datasetname: a large-scale grasping dataset including over 37 million grasps constrained to random target areas on 2889 objects.
    \item An extensive empirical evaluation of \ac{methodname} against the \sota{} 6-\ac{dof} \graspnet{}~\cite{mousavian6DOFGraspNetVariational2019c}, demonstrating, both in simulation and on real hardware, that generating constrained grasps brings significant improvement in grasp success rates and sample efficiency. 
\end{itemize}
\section{Related Work}
\label{sec:related_works}

\begin{table*}[h!]
    \centering
	\ra{1.2}\tbs{7}
	\caption{\label{tb:dataset}Comparison of constrained grasping datasets.}
\begin{tabular}{lccccc}
\toprule
           & \begin{tabular}[c]{@{}c@{}}Task-agnostic\\constraints\end{tabular} & \begin{tabular}[c]{@{}c@{}}Number of\\ objects\end{tabular} & \begin{tabular}[c]{@{}c@{}}Number of \\ grasps\end{tabular} & \begin{tabular}[c]{@{}c@{}}Grasp\\ representation\end{tabular} & \begin{tabular}[c]{@{}c@{}}Input\\ modality\end{tabular}  \\
           \midrule
Contact DB \cite{brahmbhattContactDBAnalyzingPredicting2019} & \xmark                                                           & 50                                                          & 3750                                                        & Contact map                                                    & \begin{tabular}[c]{@{}c@{}}RGB-D +\\ Thermal\end{tabular} \\
SG14000 \cite{liu2020cage}   & \xmark                                                            & 44                                                          & 14K                                                         & SE(3)                                                          & RGB-D                                                     \\
TaskGrasp \cite{muraliSameObjectDifferent2021} & \xmark                                                            & 191                                                         & 250K                                                        & SE(3)                                                          & Point cloud                                               \\
TOG-Net  \cite{fangLearningTaskorientedGrasping2020a}  & \xmark                                                            & 18K                                                         & 1.5M                                                        & SE(2)                                                          & D                                                         \\
CONG (Ours)       & \cmark                                                            & 2889                                                        & 14M                                                         & SE(3)                                                          & Point cloud \\
    \bottomrule
\end{tabular}
\end{table*}

To contextualize our work, we next review constrained grasp sampling approaches. Thereafter, we review grasping datasets for training both constrained and unconstrained data-driven grasp samplers.

\subsection{Constrained Grasp Sampling}
Early research in constrained grasping focused on analytically identifying the most appropriate task-specific grasps from already sampled grasps~\cite{borstGraspPlanningHow2004, haschkeTaskorientedQualityMeasures2005}. The central idea in those works was to formulate task-specific grasp quality metrics as optimization problems. Although the quality metrics are theoretically justifiable, calculating them requires known object models and already sampled grasps. 

More recent works \cite{detryTaskorientedGraspingSemantic2017,muraliSameObjectDifferent2021, kokicAffordanceDetectionTaskspecific2017a, songLearningTaskConstraints2010, songTaskBasedRobotGrasp2015, fangLearningTaskorientedGrasping2020a, antonovaGlobalSearchBernoulli2018a} propose data-driven methods for learning task-specific grasping to circumvent the issues with analytical methods. In one of the earliest data-driven works, \etal{Song} \cite{songLearningTaskConstraints2010,songTaskBasedRobotGrasp2015} proposed a probabilistic framework based on Bayesian networks for learning task constraints from object-action features. Another approach in \cite{fangLearningTaskorientedGrasping2020a} explored learning a classifier to determine whether a given grasp can fulfill a task.

The problem of constrained grasping can also be split into two parts: one model to detect object affordances or semantics, and another that suggests grasps on these detections \cite{detryTaskorientedGraspingSemantic2017, muraliSameObjectDifferent2021,kokicAffordanceDetectionTaskspecific2017a, liu2020cage}. For instance, in \cite{detryTaskorientedGraspingSemantic2017,kokicAffordanceDetectionTaskspecific2017a,liu2020cage}, the object affordances were first detected and then used as constraints when stochastically searching for task-specific grasps. As the iterative stochastic search process can be time-consuming, \etal{Murali} \cite{muraliSameObjectDifferent2021} proposed training a generative model to predict multiple grasps on the target object directly and then using another model to rank the proposals according to their ability to fulfill a given task. 

In this work, we draw inspiration from \cite{muraliSameObjectDifferent2021}, but instead of proposing grasps all over the target object, we learn a model that only proposes grasps on specific target areas. These areas can represent, for instance, semantically meaningful locations on the target object, such as the handle of a cup or the bottle cap, but can also cover the entire object. As such, the proposed method can generate affordance or task-oriented grasps but is not restricted to that.

\subsection{Grasping Datasets}
Due to the immense popularity and good performance of data-driven grasping methods, many different grasping datasets have been curated to train and evaluate such methods \cite{mahlerDexNetDeepLearning2017a,mousavian6DOFGraspNetVariational2019c,muraliSameObjectDifferent2021, fangGraspNet1BillionLargeScaleBenchmark2020,eppnerACRONYMLargeScaleGrasp2021a,lundellMultiFinGANGenerativeCoarseToFine2021a,depierreJacquardLargeScale2018b,leDeformationAwareDataDrivenGrasp2022, EppnerISRR2019,morrisonEGADEvolvedGrasping2020,lenzDeepLearningDetecting2015a,brahmbhattContactDBAnalyzingPredicting2019}. These datasets differ in multiple aspects, including input modality: structured \cite{mahlerDexNetDeepLearning2017a,fangGraspNet1BillionLargeScaleBenchmark2020, lundellMultiFinGANGenerativeCoarseToFine2021a, depierreJacquardLargeScale2018b, leDeformationAwareDataDrivenGrasp2022, 
morrisonEGADEvolvedGrasping2020, lenzDeepLearningDetecting2015a,brahmbhattContactDBAnalyzingPredicting2019} vs unstructured \cite{mousavian6DOFGraspNetVariational2019c, muraliSameObjectDifferent2021, eppnerACRONYMLargeScaleGrasp2021a,  EppnerISRR2019}; grasp type: planar \cite{mahlerDexNetDeepLearning2017a, depierreJacquardLargeScale2018b, leDeformationAwareDataDrivenGrasp2022, 
morrisonEGADEvolvedGrasping2020,lenzDeepLearningDetecting2015a} vs 6-\ac{dof} \cite{mousavian6DOFGraspNetVariational2019c, muraliSameObjectDifferent2021,    fangGraspNet1BillionLargeScaleBenchmark2020,eppnerACRONYMLargeScaleGrasp2021a,lundellMultiFinGANGenerativeCoarseToFine2021a,EppnerISRR2019,brahmbhattContactDBAnalyzingPredicting2019}; and labels: simulation \cite{mousavian6DOFGraspNetVariational2019c, muraliSameObjectDifferent2021, eppnerACRONYMLargeScaleGrasp2021a, leDeformationAwareDataDrivenGrasp2022, EppnerISRR2019} vs analytic \cite{mahlerDexNetDeepLearning2017a,fangGraspNet1BillionLargeScaleBenchmark2020, lundellMultiFinGANGenerativeCoarseToFine2021a,morrisonEGADEvolvedGrasping2020} vs human annotated \cite{lenzDeepLearningDetecting2015a,brahmbhattContactDBAnalyzingPredicting2019}.    

Despite the abundance of grasping datasets, only a few exist for constrained grasping, as highlighted in \tabref{tb:dataset}. Most of these constrained grasping datasets \cite{brahmbhattContactDBAnalyzingPredicting2019,liu2020cage,muraliSameObjectDifferent2021}, are relatively small, including, at most, 191 objects and 250,000 grasps. Moreover, the datasets in \cite{brahmbhattContactDBAnalyzingPredicting2019,liu2020cage,muraliSameObjectDifferent2021} all require humans to create or label the grasps. However, one of the most pressing limitations of all the previously proposed constrained grasping datasets \cite{brahmbhattContactDBAnalyzingPredicting2019,liu2020cage,muraliSameObjectDifferent2021,fangGraspNet1BillionLargeScaleBenchmark2020} is that the grasps are conditioned on specific tasks, rendering them unusable for training constrained grasping policies that can generate grasps on \textit{any} constrained area. In this work, we propose a new dataset called \datasetname{}, inspired from \cite{muraliSameObjectDifferent2021} but did neither require human labeling nor grasps tied to specific tasks. \datasetname{}, with over 14 million training examples on 2889 objects, is also orders of magnitude larger compared to \cite{brahmbhattContactDBAnalyzingPredicting2019,liu2020cage,muraliSameObjectDifferent2021}.
\section{Problem Statement}
\label{sec:problem_statement}

\begin{figure}[t!]
  \begin{center}
    \includegraphics[width=0.18\textwidth]{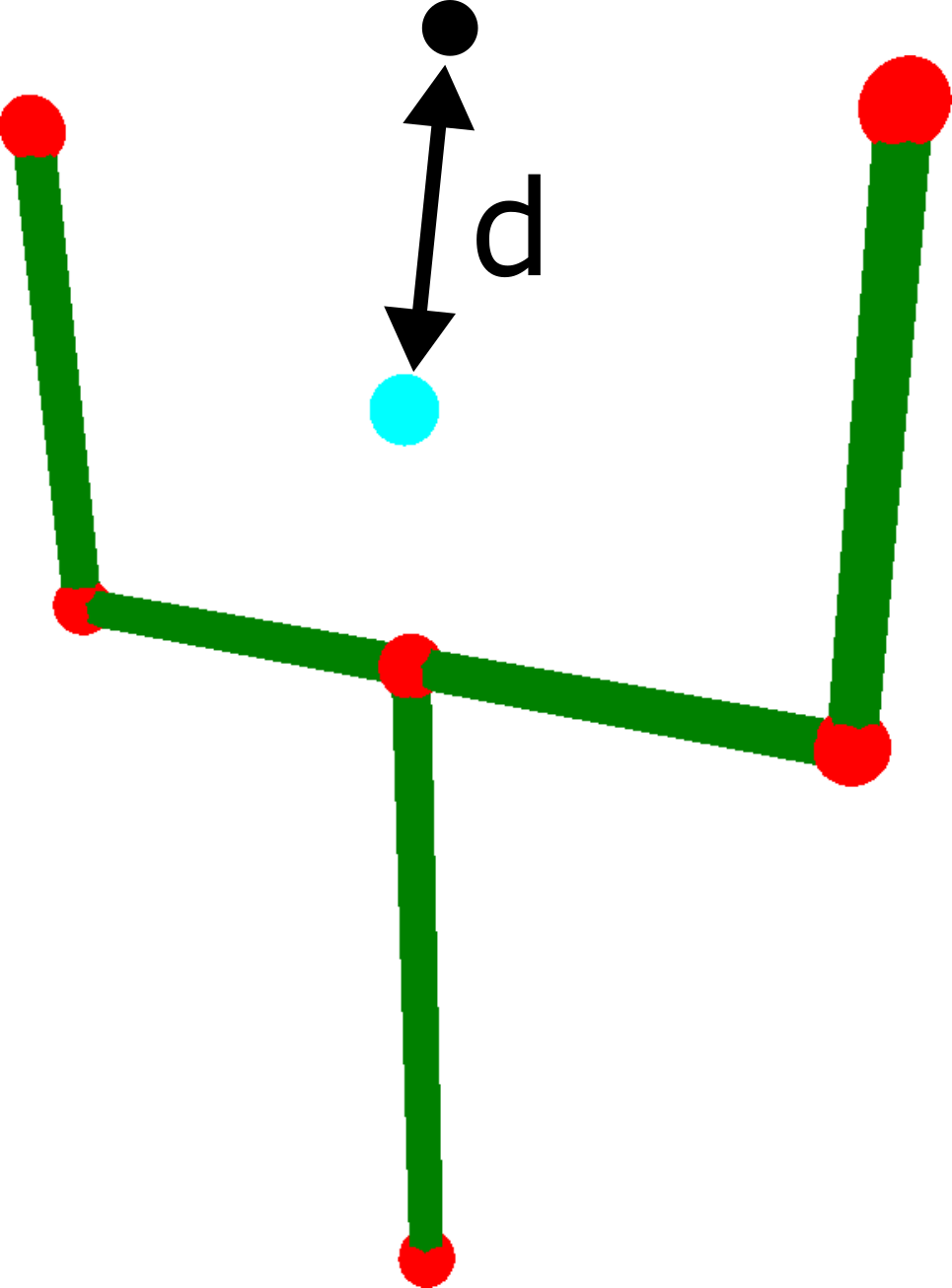}
  \end{center}
  \caption{The grasp in green, its \pc{} representation in red, the center grasp point in cyan, and the distance d between it and the black point on the object. The center grasp point is set to the average of the two leftmost and the two rightmost points of the gripper.  \label{fig:grasp_point}}
\end{figure}

In this work, we address the problem of generating parallel-jaw grasp poses $\matr{G}$ on an object point-cloud $\matr{O}\in \mathbb{R}^{\text{N}\times 3}$ such that when the gripper is closed, the grasps are both stable ($\text{S}=1$) and located at a target area $\matr{A}\subseteq \matr{O} \in \mathbb{R}^{\text{M}\times 3}$.
Here, $\text{N}$ and $\text{M}$, where $\text{M}\leq \text{N}$, represent the number of points in a \pc{}. As presented in \figref{fig:grasp_point}, a grasp $\matr{G}$ is located at a target area $\matr{A}$ iff the grasp center point is at most at a Euclidean distance d from any point in $\matr{O}$.

Specifically, the target is to learn the joint distribution $\prob{(\matr{G},\text{S}\mid\matr{O},\matr{A})}$.  To learn such a complex joint distribution, we propose factorizing it into two separate distributions:
\begin{enumerate*}[label=\arabic*)]
    \item a generative grasp sampler $\prob{(\matr{G}\mid\matr{O},\matr{A})}$ from which constrained grasps can be sampled, and
    \item a grasp evaluator $\prob{(\text{S}\mid\matr{O},\matr{G})}$ for evaluating the stability of each grasp $\matr{G}$.
\end{enumerate*}
We approximate these distributions with parametric models $\mathcal{Q}_{\mathbf{\theta}}(\matr{G}\mid\matr{O},\matr{A})\approx \prob{(\matr{G}\mid\matr{O},\matr{A})}$ and $\mathcal{M}_{\mathbf{\phi}}(\text{S}\mid\matr{O},\matr{G})\approx \prob{(\text{S}\mid\matr{O},\matr{G})}$, where $\boldsymbol{\theta}$ and $\boldsymbol{\phi}$ are trainable parameters. 

We represent a grasp $\matr{G}=[\mathbf{q},~\mathbf{p}]$ by a unit quaternion $\mathbf{q}\in \mathbb{R}^4$ and a 3-D position $\mathbf{p}\in \mathbb{R}^3$. Because we use quaternions, the grasp pose is represented by 7 scalars. We assume that all sampled grasps are reachable and that only one graspable object is present.

Solving the aforementioned problem requires the following: 
\begin{enumerate*}[label=(\roman*)]
    \item the parametric models $\mathcal{Q}_{\mathbf{\theta}}$ and $\mathcal{M}_{\mathbf{\phi}}$ capable of approximating their target distributions, and \item a dataset for training the parametric models.
\end{enumerate*}  
In the next section, the parametric models are presented, and in \secref{sec:dataset}, the dataset used for training them is described. 

\section{Method}
\label{sec:method}

In the previous section, the problem of learning to sample stable constrained grasps was separated into learning a parametric grasp sampler $\mathcal{Q}_{\boldsymbol{\theta}}$ and a parametric grasp evaluator $\mathcal{M}_{\boldsymbol{\phi}}$. In this section, we first present the constrained grasp sampler (\secref{sec:constrained_grasp_sampler}) and then the grasp evaluator (\secref{sec:grasp_evaluator}).

\subsection{Variational Constrained Grasp Sampler}
\label{sec:constrained_grasp_sampler}
We model the parametric grasp sampler $\mathcal{Q}_{\mathbf{\theta}}(\matr{G}\mid\matr{O},\matr{A})$ using a \ac{cvae}~\cite{sohnLearningStructuredOutput2015}, where the conditional variables are $\matr{O}$ and $\matr{A}$. The parametric grasp sampler, henceforth referred to as \ac{methodname}, is the decoder $\mathsf{p}_{\boldsymbol{\chi}}(\matr{G}\mid\matr{O},\matr{A},\mathbf{z})$ of the \ac{cvae}. The corresponding encoder $\mathsf{q}_{\boldsymbol{\psi}}(\mathbf{z}\mid\matr{O}, \matr{A}, \matr{G})$ is only used to train the encoder-decoder network using examples of high-quality grasps. 
$\boldsymbol{\psi}$ and $\boldsymbol{\chi}$ represent the trainable parameters and $\mathbf{z}\in \mathbb{R}^{\text{L}}$ is a latent space variable of size $\text{L}$. 

The backbone for both the encoder and decoder is PointNet++~\cite{qiPointNetDeepHierarchical2017a}. Because of this choice, the  network input has to be in the form of a \pc{} $\matr{X} \in \mathbb{R}^{\text{N}\times (3+\text{K})}$, where each point $\mathbf{x} \in \matr{X}$ is represented by its 3D Euclidean position and, optionally, $\text{K}$ additional real-valued or binary features. For both the encoder and decoder, $\matr{X}$ is the same as $\matr{O}$ but with an additional point-wise binary feature indicating if the point $\mathbf{x} \in \matr{X}$ belongs to the target area $\matr{A}$ or not. This construction is made possible because, as defined in \secref{sec:problem_statement}, $\matr{A}\subseteq \matr{O}$. 

In addition to the extra binary input feature, the encoder also takes $\mathbf{g}$ as a point-wise feature. The input dimension then becomes $\text{N} \times 11$, where $\text{N}$ is the number of points in the \pc{}, and the eleven features consist of the 3D position of each point, the binary feature indicating if the point belongs to the target area or not, and the 7-dimensional grasp pose representation. The decoder, on the other hand, does not include $\mathbf{g}$ but does include the latent space variable  $\mathbf{z}\in \mathbb{R}^{\text{L}}$ as an additional point-wise feature. Therefore, the decoder input dimension is $\text{N} \times (4+\text{L})$, where $\text{L}$ is the dimension of the latent space. In this work, the size of the latent space was set to 2 in accordance with prior work \cite{mousavian6DOFGraspNetVariational2019c}. 

\ac{methodname} is trained on the standard \ac{elbo} loss:
\begin{align}
\label{eq:vae_loss}
    \mathcal{L}_{\text{VAE}} = \mathcal{L}(\matr{G}^*,\hat{\matr{G}}) + \alpha \mathcal{D}_{\text{KL}}[\mathsf{q}_{\psi} (\mathbf{z}\mid\matr{X},\matr{G}^*),~\mathcal{N}(\matr{0},\matr{I})],
\end{align}
where $\alpha$ is a scalar, and $\mathcal{D}_{\text{KL}}$ is the KL-divergence between the latent space encoding $\mathbf{z}$ produced by $\mathsf{q}_{
\psi}$, and a zero-mean Gaussian distribution. The reconstruction loss $\mathcal{L}$ in \eqref{eq:vae_loss} is defined as
\begin{align}
    \mathcal{L}(\matr{G}^*,\hat{\matr{G}}) = \norm{\text{h}(\matr{G}^*)-\text{h}(\mathbf{\hat{\matr{G}}})}_1,
\end{align}
where $\matr{G}^*$ is a ground truth stable grasp, $\matr{\hat{G}}$ the generated grasp from the decoder $\mathsf{p}_{\chi}$, and $\text{h}: \mathbb{R}^{7} \rightarrow  \mathbb{R}^{6\times 3}$ is a function that maps a 7D grasp pose into a \pc{} representation of the gripper $\matr{P} \in \mathbb{R}^{6\times 3}$ as visualized in \figref{fig:grasp_point}. The reason for mapping grasp poses to \pcs{} is that it combines both the translation and orientation components into a single loss function~\cite{mousavian6DOFGraspNetVariational2019c}.  

Both the encoder and the decoder are optimized during training, while only the decoder is used for grasp sampling. More specifically, to sample a set of grasps $\matr{\hat{G}}$, the first step is to draw multiple random latent samples $\mathbf{z}\sim \mathcal{N}(\matr{0},\matr{I})$. Then, each of these samples is concatenated to a separate copy of the input \pc{} together with the binary feature representing the target area. Finally, each copy of the \pc{} is passed through the decoder, producing a separate grasp.

\begin{figure*}[t!]
\centering
\begin{subfigure}[t]{.22\textwidth}
  \centering
  \includegraphics[height=0.9\linewidth, keepaspectratio,width=1\linewidth]{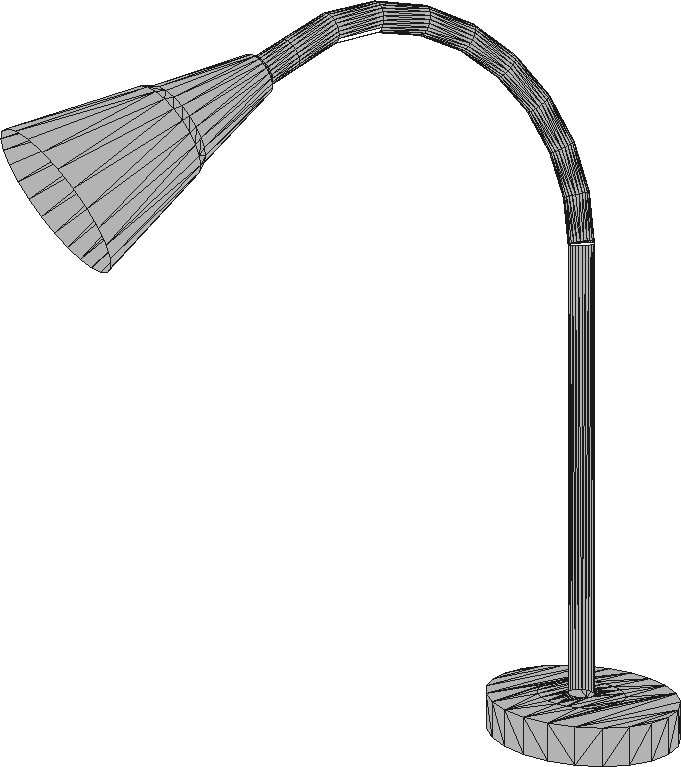}
  \caption{}
  \label{fig:sub1}
\end{subfigure}%
\begin{subfigure}[t]{.22\textwidth}
  \centering
   \includegraphics[height=0.9\linewidth, keepaspectratio,width=1\linewidth]{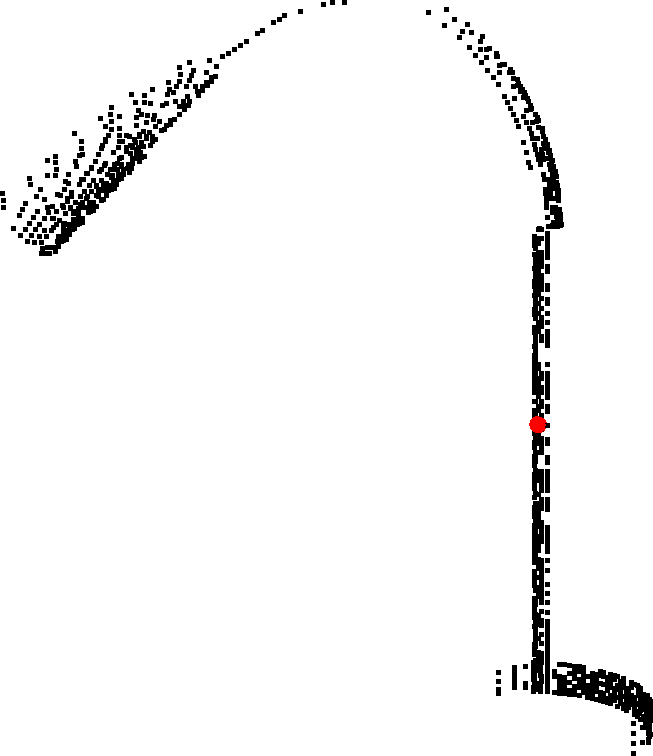}
  \caption{}
  \label{fig:sub2}
\end{subfigure}
\begin{subfigure}[t]{.22\textwidth}
  \centering
   \includegraphics[height=0.9\linewidth, keepaspectratio,width=1\linewidth]{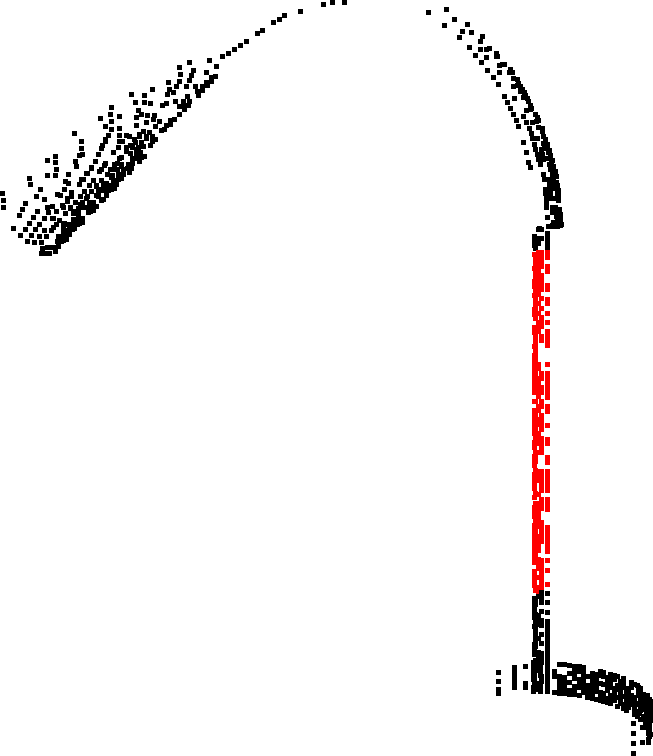}
  \caption{}
  \label{fig:sub3}
\end{subfigure}
\begin{subfigure}[t]{.30\textwidth}
  \centering
   \includegraphics[height=0.7\linewidth, keepaspectratio,width=1\linewidth]{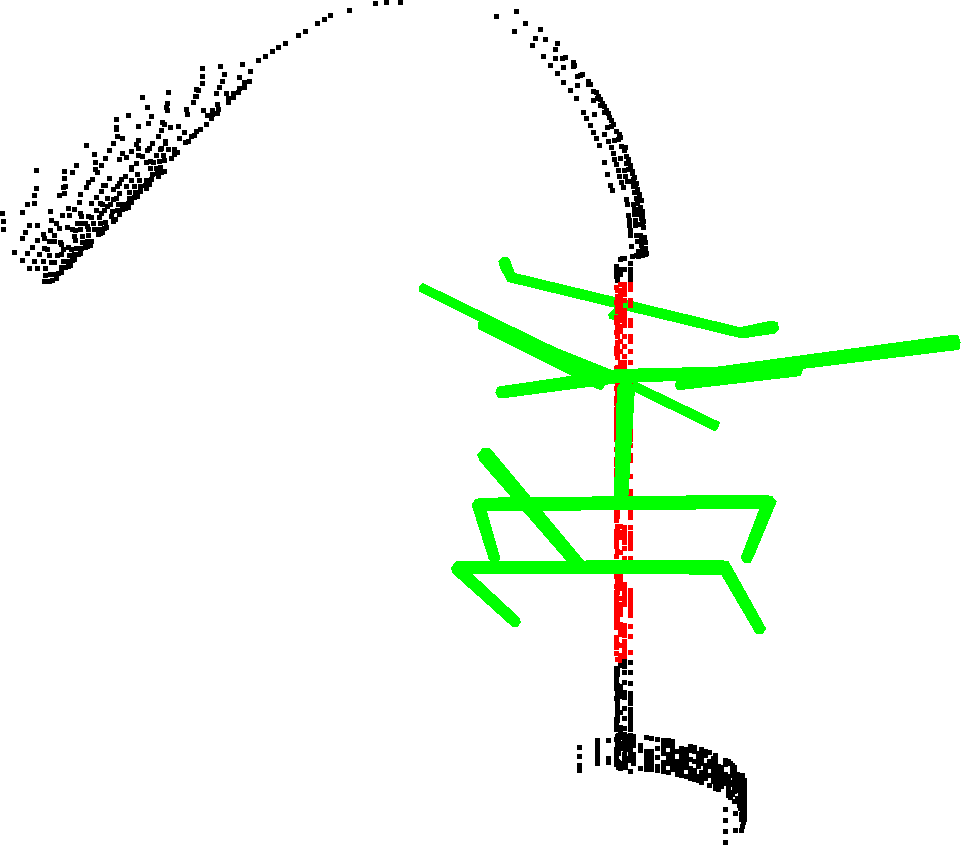}
  \caption{}
  \label{fig:sub4}
\end{subfigure}
\caption{An example of how the dataset is curated. (a) From the object mesh, (b) a point cloud is rendered, and a query point, highlighted in red, is selected. Given the query point, (c) all neighbors within a specific radius from it are found, and (d) the grasps close to those points are stored.}
\label{fig:dataset_generation}
\end{figure*}

\subsection{Grasp Evaluator}
\label{sec:grasp_evaluator}

Because the constrained grasp sampler is only trained on stable grasps, it can learn to generate poor grasps between modes~\cite{mousavian6DOFGraspNetVariational2019c}. To avoid executing poor grasps, we train a grasp evaluator to distinguish between good and bad grasps by predicting the probability that a grasp $\matr{G}$ succeeds ($\text{S}=1$) on object $\matr{O}$. Formally, the grasp evaluator models the conditional probability $\prob{(\text{S}\mid\matr{G},\matr{O})}$.

The grasp evaluator used in this work was originally proposed in \cite{mousavian6DOFGraspNetVariational2019c}. It is formed as a deep network that uses the PointNet++ architecture \cite{qiPointNetDeepHierarchical2017a} as the backbone. Therefore, the input to the evaluator network is also a \pc{} $\matr{Y} \in \mathbb{R}^{(\text{N}+6)\times (3+1)}$. However, in contrast to the input \pc{} to \ac{methodname}, $\matr{Y}$ consists of an object \pc{} $\matr{O} \in \mathbb{R}^{\text{N}\times 3}$ concatenated with a grasp pose \pc{} $\matr{P} \in \mathbb{R}^{6\times 3}$, and an additional point-wise binary feature to distinguish between these two \pcs. 

The grasp evaluator network is trained on the binary cross-entropy loss
\begin{align}
\mathcal{L}_{E}=-\text{S}^*\log(\text{S})+(1-\text{S}^*)\log(1-\text{S}),    
\end{align}
where $\text{S}^*$ is the ground-truth success of a grasp and $\text{S}$ is the predicted success.

\section{Dataset}
\label{sec:dataset}

To train \ac{methodname}, we need a large-scale dataset consisting of object \pcs{} $\matr{O}$, and successful grasps $\matr{G}^*$ on randomly sampled target areas $\matr{A}$. To curate such a dataset, we expand the recently large-scale Acronym dataset \cite{eppnerACRONYMLargeScaleGrasp2021a} to include randomly subsampled grasping areas. We name the new dataset \datasetname{}. 

An overview of the process to curate \datasetname{} is presented in \figref{fig:dataset_generation}. Formally, the process is divided into four steps: 
\begin{enumerate}[label=(\roman*)]
    \item Place the object at the origin in a randomized orientation and render a \pc{} $\matr{O}\in \mathbb{R}^{\text{N}\times 3}$ from it.
    \item Sample $\matr{I}\in \mathbb{R}^{\text{K}\times 3}$ query points from $\matr{O}$, where $\text{K}\ll \text{N}$, using the \ac{fps} algorithm.
    \item For each query point $\mathbf{x}_i\in \matr{I}$, find all neighboring points $\matr{A}_i$ that are within a uniformly sampled radius $r_i\sim \mathcal{U} [0,~\text{R}]$ from $\mathbf{x}_i$, where $\text{R}$ is the diagonal length of the mesh's bounding box.
    \item  \label{enum:step3} Find all grasps $\matr{G}$ for object $\matr{O}$ in the Acronym dataset \cite{eppnerACRONYMLargeScaleGrasp2021a} where the distance between the center grasp point and any point in $\matr{A}_i$ is at most d.    
\end{enumerate}  
For the steps above, we defined $\text{N} = 1024$, $\text{K}= 50$, and $\text{d} = \SI{2}{\cm}$. The center grasp point is defined as in \figref{fig:grasp_point}. 

We ran the above procedure on 2889 objects from the Acronym dataset~\cite{eppnerACRONYMLargeScaleGrasp2021a}. For each of the 2889 objects, we rendered 100 \pcs{} $\matr{O}$ of the object, and for each of these \pcs{}, we sampled $\matr{I}$. The resulting dataset contains over 14 million examples with an average of 257 grasps per target area $\matr{A}_i$. Of this dataset, 123 objects were randomly selected for the simulated grasping experiment, and the rest were used for training.
\section{Experiments}
\label{sec:experiments}

The two questions we want to answer in the experiments are:
\begin{enumerate}
    \item What is the grasp success rate of constrained grasping?
    \item How much more sample efficient is a constrained grasp sampler than an unconstrained one for target-driven grasping?
\end{enumerate}

We answer these two questions with two experiments: one in simulation and one using real robotic hardware. In all experiments, \ac{methodname} was benchmarked against \graspnet~\cite{mousavian6DOFGraspNetVariational2019c}. Both methods were trained on the same objects from Acronym~\cite{eppnerACRONYMLargeScaleGrasp2021a} to ensure a just comparison. 

To evaluate grasps, we used the grasp success rate metric, which is the ratio of successful grasps to total grasp attempts. We counted a grasp as successful if the object was successfully picked up and remained within the gripper during a predefined manipulation motion. To only test grasps on the target area, all those whose center grasping point was further away than a distance $d$ to any point in the target area were removed. As in \secref{sec:dataset}, we set $d = \SI{2}{\cm}$. 

\subsection{Simulated Robotic Grasping}
\label{sec:simulation}

\begin{table*}[h!]
    \centering
	\ra{1.7}\tbs{5}
	\caption{\label{tb:results_simulation}Simulation results when evaluating the 10 highest scoring grasps. In cases where less than 10 grasps were kept, all were evaluated. $\uparrow$: higher the better, $\downarrow$: lower the better.}
\scalebox{1.0}{
\begin{tabular}{lcccccccccccccccc}
\toprule

                                            & \multicolumn{6}{c}{Unconstrained sampling}                                                                  & \multicolumn{10}{c}{Constrained sampling}   
                                            \\
                                            \cmidrule(lr){2-7}\cmidrule(lr){8-17}
                                            & \multicolumn{2}{c}{\ac{methodname}}     & \multicolumn{4}{c}{\graspnet{}} & \multicolumn{2}{c}{\ac{methodname}}    & \multicolumn{4}{c}{\graspnet{}} & \multicolumn{4}{c}{\graspnet \acs{tai}{}}            \\
                                            \cmidrule(lr){2-3}\cmidrule(lr){4-7}\cmidrule(lr){8-9}\cmidrule(lr){10-13}\cmidrule(lr){14-17}

\# of grasps sampled                        & 100                           & 500                      & 100   & 500   & 5K                       & 10K   & 100                          & 500                      & 100   & 500   & 5K                       & 10K   & \multicolumn{1}{c}{100} & \multicolumn{1}{c}{500} & 5K & 10K  \\
\midrule
Grasp success rate (\%)~$\uparrow$     & \textbf{56} & \textbf{56}                       & 45    & 45    & 47                       & 45    & 53 & \textbf{54}                       & 39    & 43    & 46                       & 46    & 34                      & 33                      & 33   & 34   \\
Ratio of grasps kept (\%)~$\uparrow$ & 100                           & 100                      & 100   & 100   & 100                      & 100   & \textbf{93} & \textbf{93}                       & 30    & 30    & 30                       & 30    & 34                      & 34                      &  34  & 34   \\
Inference time (s)~$\downarrow$        & \textbf{0.07}                          & 0.29 & \textbf{0.07}  & 0.31  & 5.63 & 10.27 & 0.07                         & 0.27 & 0.07  & 0.27  & 3.9  & 8.55  & \textbf{0.06} & 0.27 &  3.93  & 8.11 \\
\bottomrule
\end{tabular}
}
\end{table*}

\begin{wrapfigure}[14]{r}{0.14\textwidth}
  \begin{center}
    \includegraphics[width=0.14\textwidth]{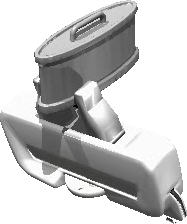}
  \end{center}
  \caption{An example grasp from the simulation.\label{fig:grasp_in_sim}}
\end{wrapfigure}

In the simulation experiments, we wanted to test the best grasp, not the best reachable one. Therefore, to ensure all sampled grasps were reachable, we used a free-floating \franka{} gripper to grasp a free-floating object, as depicted in \figref{fig:grasp_in_sim}. 
To grasp an object, an open \franka{} gripper was placed at the grasp pose, and then the finger closed slowly until either the object was grasped or the two fingers touched. If the object was between the fingers of the gripper, the grasp was evaluated by turning gravity on and executing a predefined linear acceleration motion followed by an angular acceleration motion. The grasp was successful if the object remained within the gripper during all motions. 

The simulation experiments were carried out using the publicly available \isaacgym{} simulator~\cite{makoviychukIsaacGymHigh2021a} on 123 randomly held out objects from the Acronym dataset \cite{eppnerACRONYMLargeScaleGrasp2021a}. To observe the objects, we used a simulated depth sensor. 

Two different simulation experiments were conducted to determine the effect constrained grasp sampling had on grasp success rates. In the first experiment, called \emph{Unconstrained sampling}, grasps around the objects were sampled, and no specific target area was set. To sample unconstrained grasps with \ac{methodname}, all the points were set as the target area\ie{} $\matr{A} = \matr{O}$. In the second simulation experiment, called \emph{Constrained sampling}, only grasps at the target area were allowed. We also included an additional baseline in the second experiment called \graspnetta{} that used only the target area as input to the grasp sampling network. 

To analyze the effect constrained sampling had on sample efficiency, we varied the number of grasps sampled. For \ac{methodname}, we sampled 100 and 500 grasps per object or target area, while for \graspnet{}, we sampled 100, 500, 5000, and 10000 grasps. Out of these grasps, we only executed the 10 highest-scoring grasps according to the grasp evaluator. The same procedure as in \secref{sec:dataset} was used for generating the target areas. That is, from the rendered \pc{}, we first sampled 10 query points using \ac{fps}. Then, for each query point, a target area was constructed by finding all neighboring points within a uniformly sampled radius from the query point. The bounds on the uniform distribution were $0$ and $\text{R}$, where $\text{R}$ is the diagonal length of the mesh's bounding box. 

The experimental results are presented in \tabref{tb:results_simulation}, and an example grasp is shown in \figref{fig:grasp_in_sim}. Based on these results, we can draw multiple conclusions. First, for constrained grasping, \ac{methodname} kept over three times more grasps than \graspnet{}, demonstrating the benefit of constraining grasp sampling already in the input to the network. Moreover, \graspnetta{} did not generate more successful grasps than \graspnet{}, highlighting the benefit of using global object information even when sampling grasps locally at specific target regions. 

Secondly, for \graspnet{}, the number of sampled grasps mainly affects the grasp success rate for constrained sampling, but the effect tapers with the number of grasps. We hypothesize that this finding demonstrates that an unconstrained grasp sampler must sample orders of magnitude more grasps to increase the probability that some good ones end up in the target area. This hypothesis is also supported by the fact that for the same experiment, the ratio of grasps kept for \graspnet{} is the same no matter if 100, 500, 5000, or 10000 grasps were sampled. However, by sampling orders of magnitude more grasps, as in the case of 5000 or 10000 grasps, the inference time increases by 80--140 times compared to only sampling 100 grasps.  

The last conclusion to draw is that \ac{methodname} achieves the highest grasp success rates in both constrained and, more interestingly, unconstrained grasping. We hypothesize that the lower unconstrained grasp success rate for \graspnet{} is because it generates many more grasps on the unobserved part of the object than \ac{methodname}. The reason for generating grasps on unobserved parts is that these could admit better grasps than the observed parts. However, the grasp success prediction for such grasps is also less reliable and could lead to more misclassified grasps.  

All in all, the results from the simulation experiment highlight the benefits of constraining grasp sampling. Next, we investigate if similar benefits are present in real-world robotic grasping.  

\subsection{Real Robotic Grasping}
\label{sec:real}

\begin{figure*}[h!]
\centering
\begin{subfigure}[t]{.15\textwidth}
  \centering
  \scalebox{-1}[1]{\includegraphics[height=0.7\linewidth, keepaspectratio,width=1\linewidth]{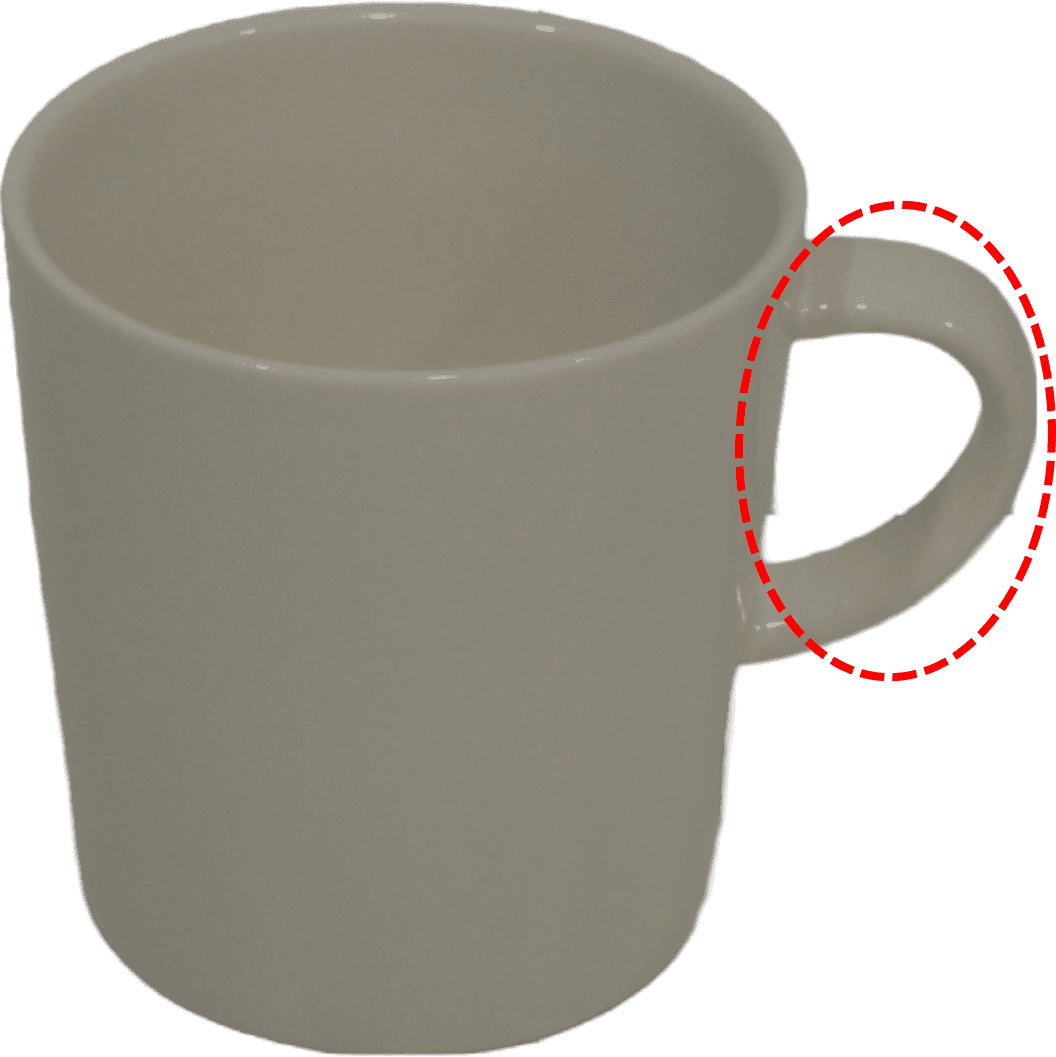}}
  \caption{}
  \label{fig:grasp_cup}
\end{subfigure}%
\begin{subfigure}[t]{.15\textwidth}
  \centering
  \includegraphics[height=0.7\linewidth, keepaspectratio,width=1\linewidth]{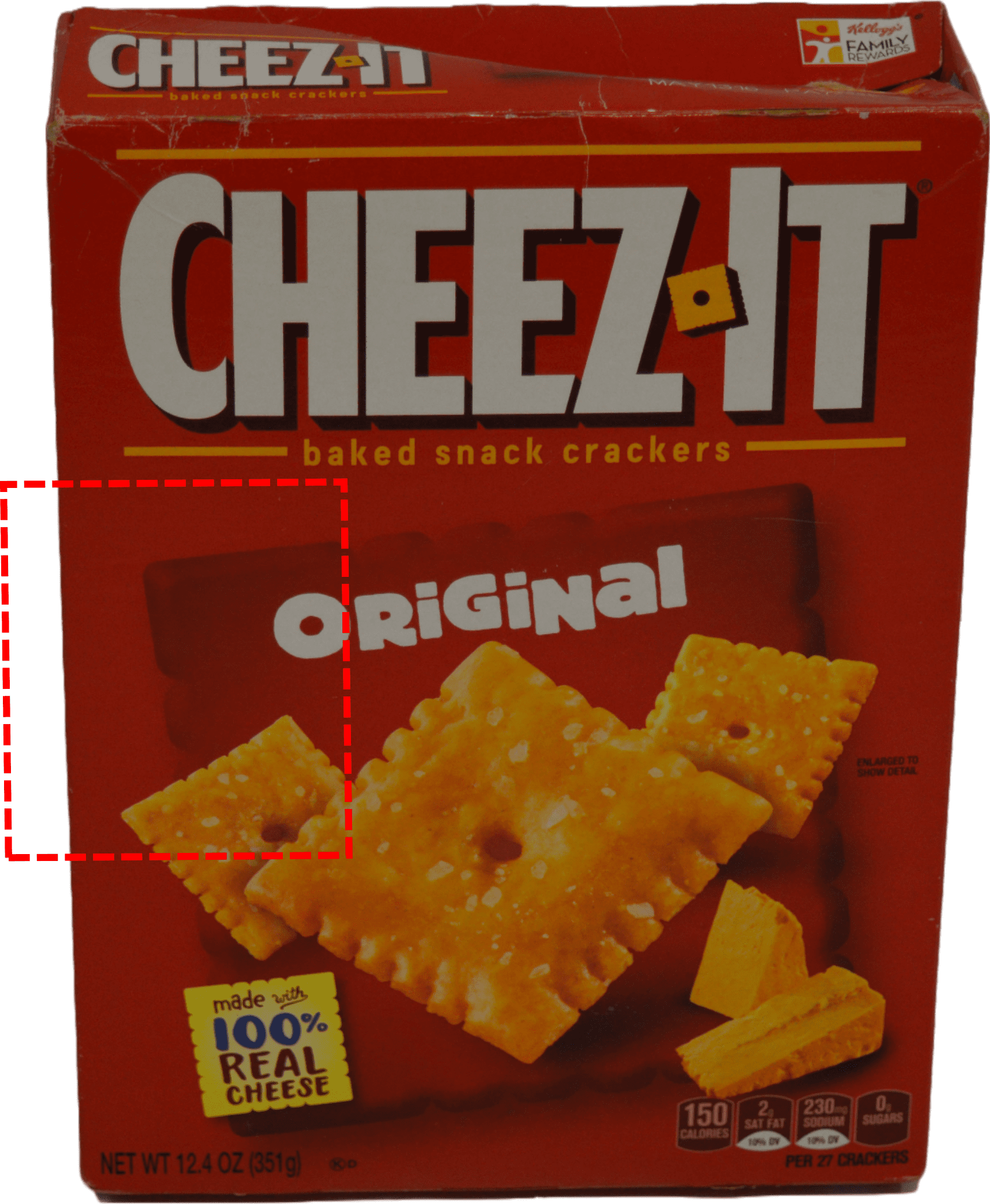}
  \caption{}
  \label{fig:grasp_cheezeit}
\end{subfigure}%
\begin{subfigure}[t]{.15\textwidth}
  \centering
  \includegraphics[height=0.7\linewidth, keepaspectratio,width=1\linewidth]{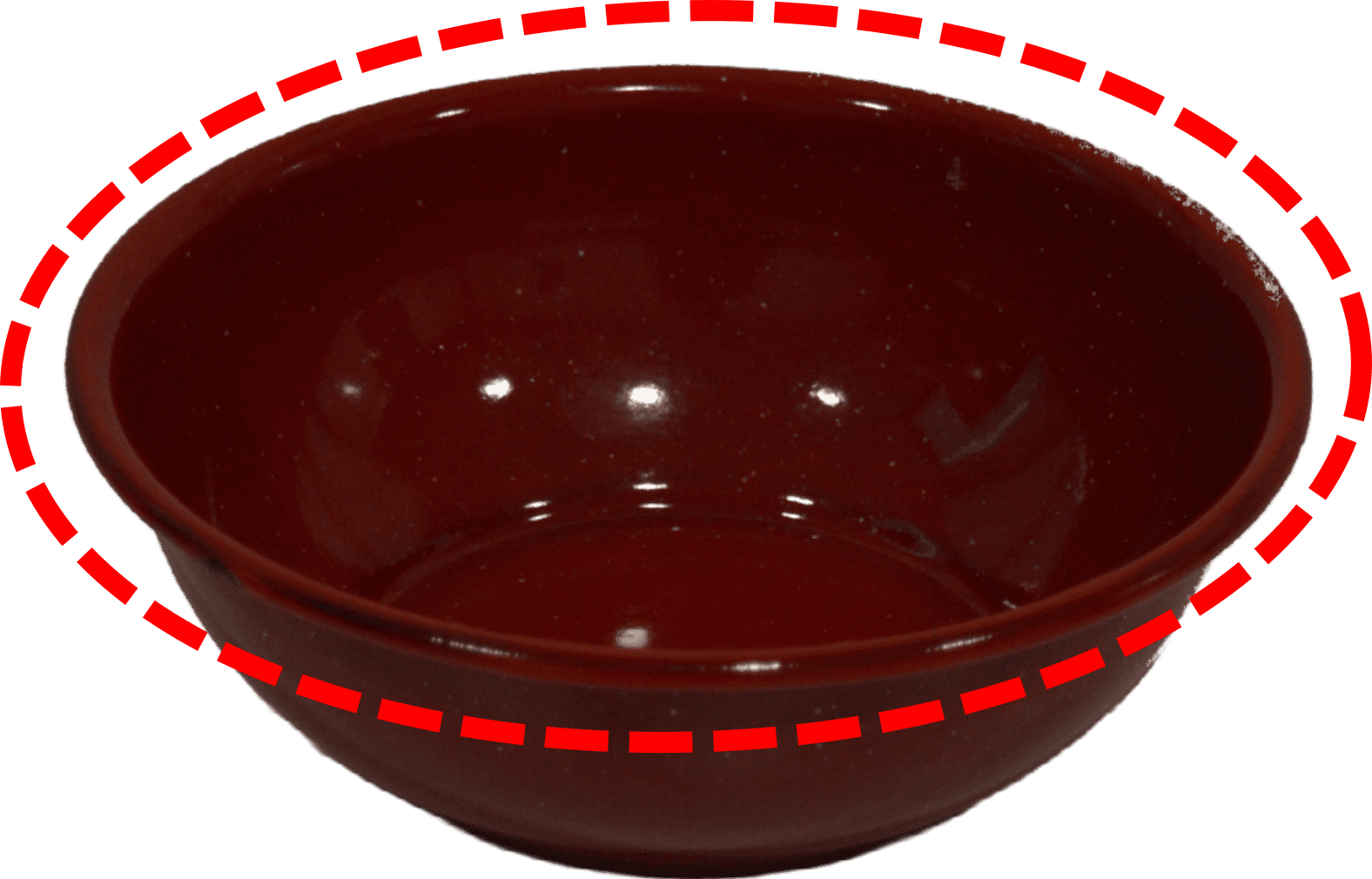}
  \caption{}
  \label{fig:grasp_bowl}
\end{subfigure}%
\begin{subfigure}[t]{.15\textwidth}
  \centering
  \includegraphics[height=0.7\linewidth, keepaspectratio,width=0.6\linewidth]{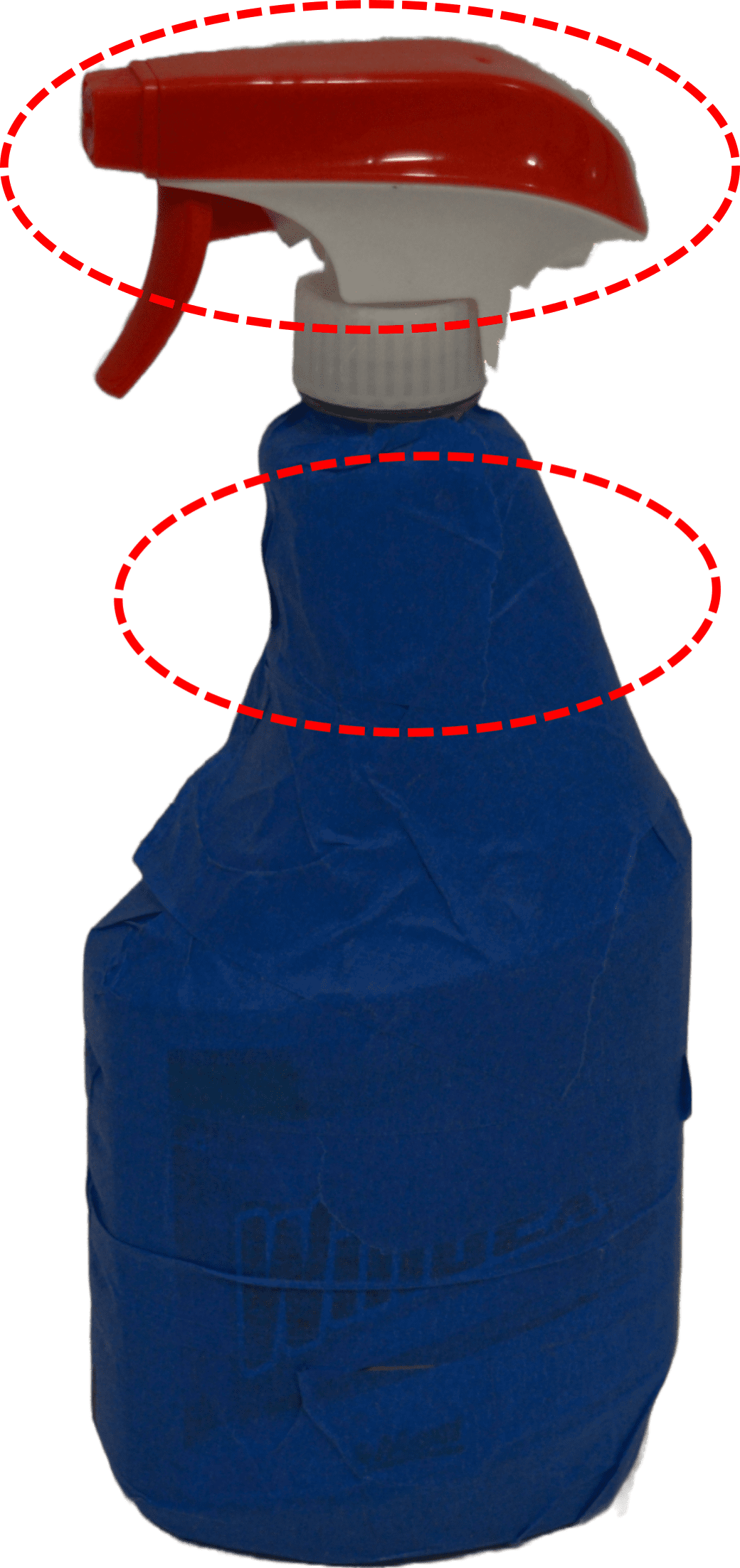}
  \caption{}
  \label{fig:grasp_spray}
\end{subfigure}%
\begin{subfigure}[t]{.15\textwidth}
  \centering
  \includegraphics[height=0.7\linewidth, keepaspectratio,width=0.9\linewidth]{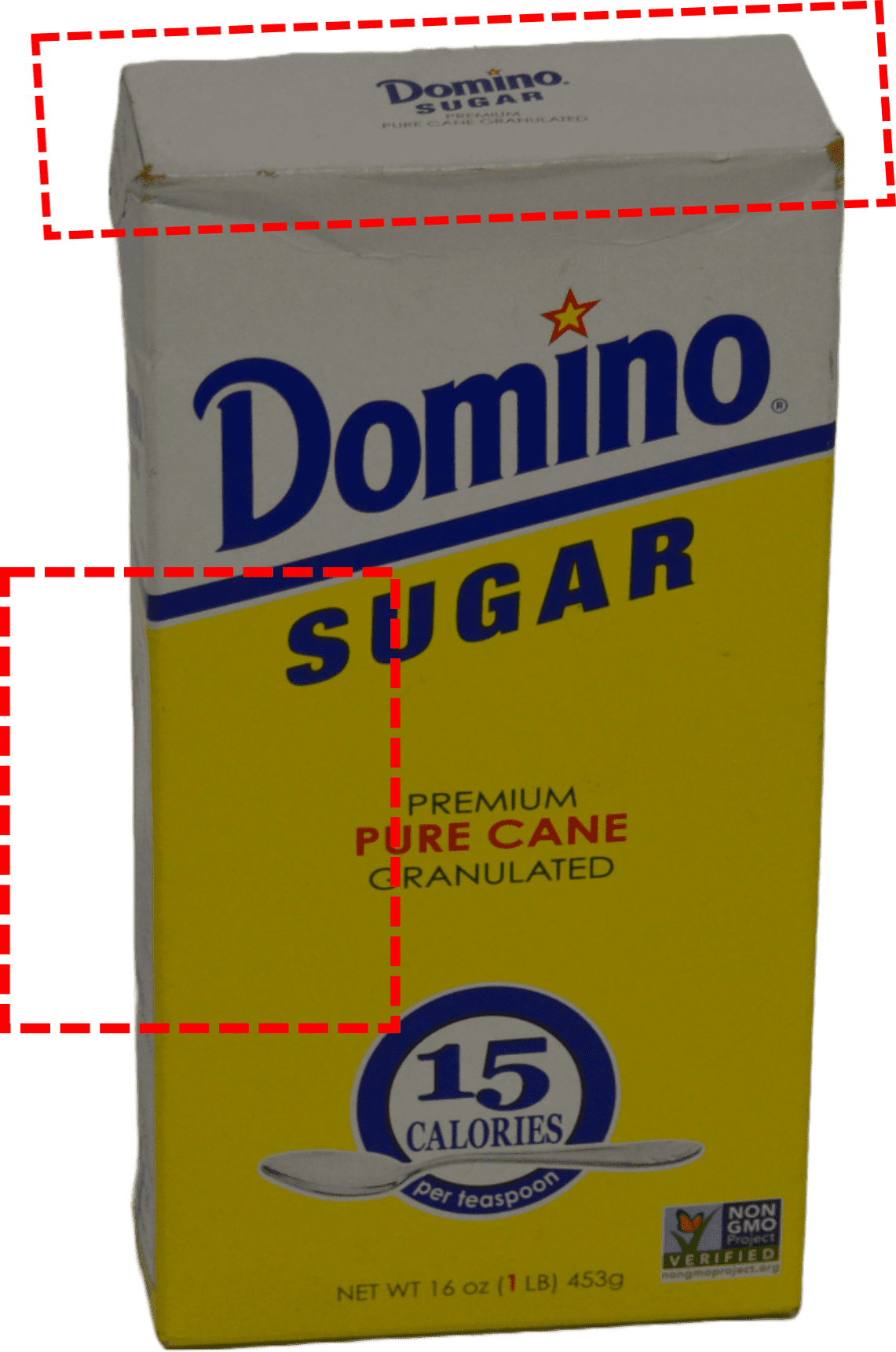}
  \caption{}
  \label{fig:grasp_domino}
\end{subfigure}%
\begin{subfigure}[t]{.15\textwidth}
  \centering
  \includegraphics[height=0.7\linewidth, keepaspectratio,width=1\linewidth]{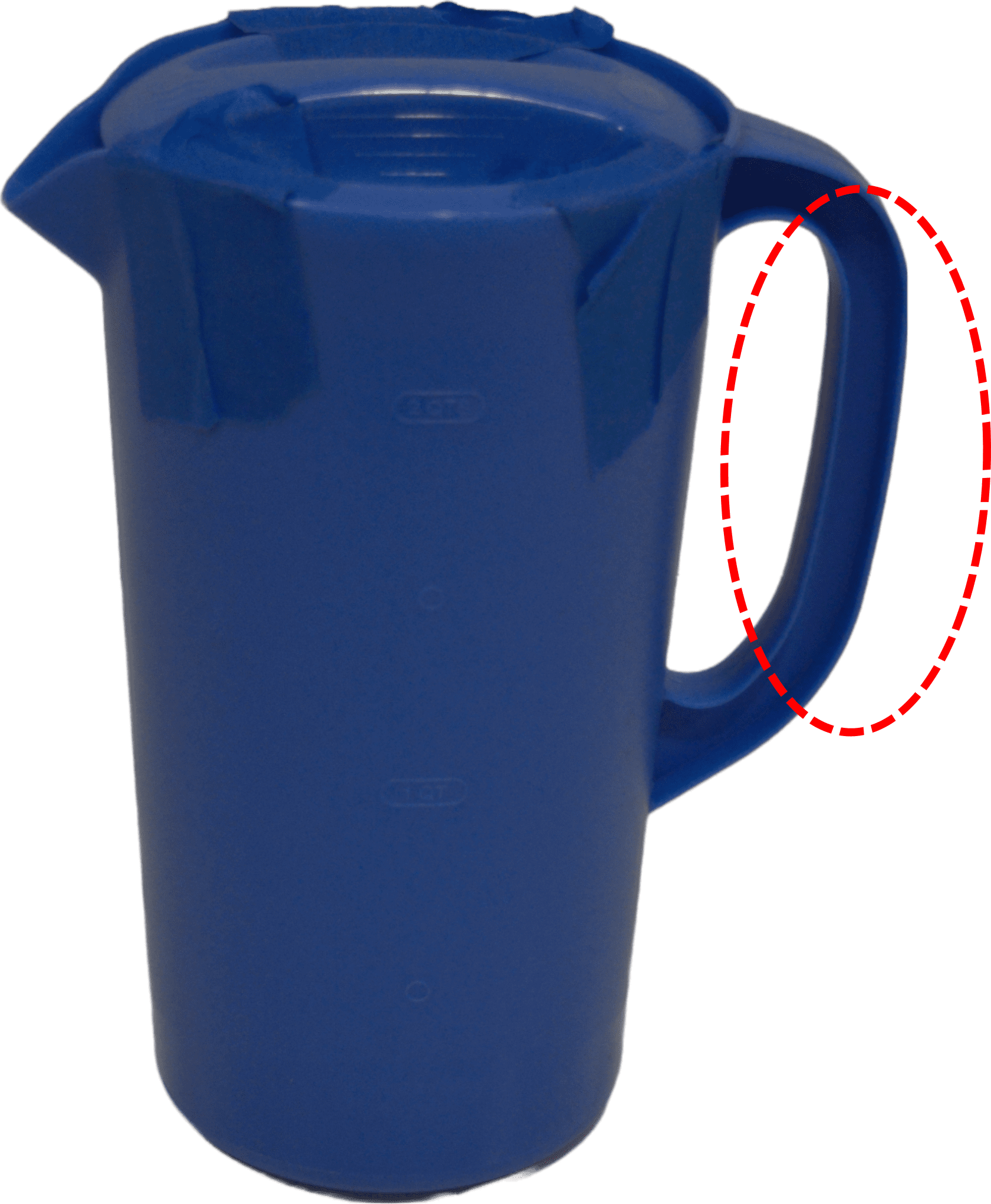}
  \caption{}
  \label{fig:grasp_pitcher}
\end{subfigure}\\
\begin{subfigure}[t]{.15\textwidth}
  \centering
  \includegraphics[height=0.7\linewidth, keepaspectratio,width=1\linewidth]{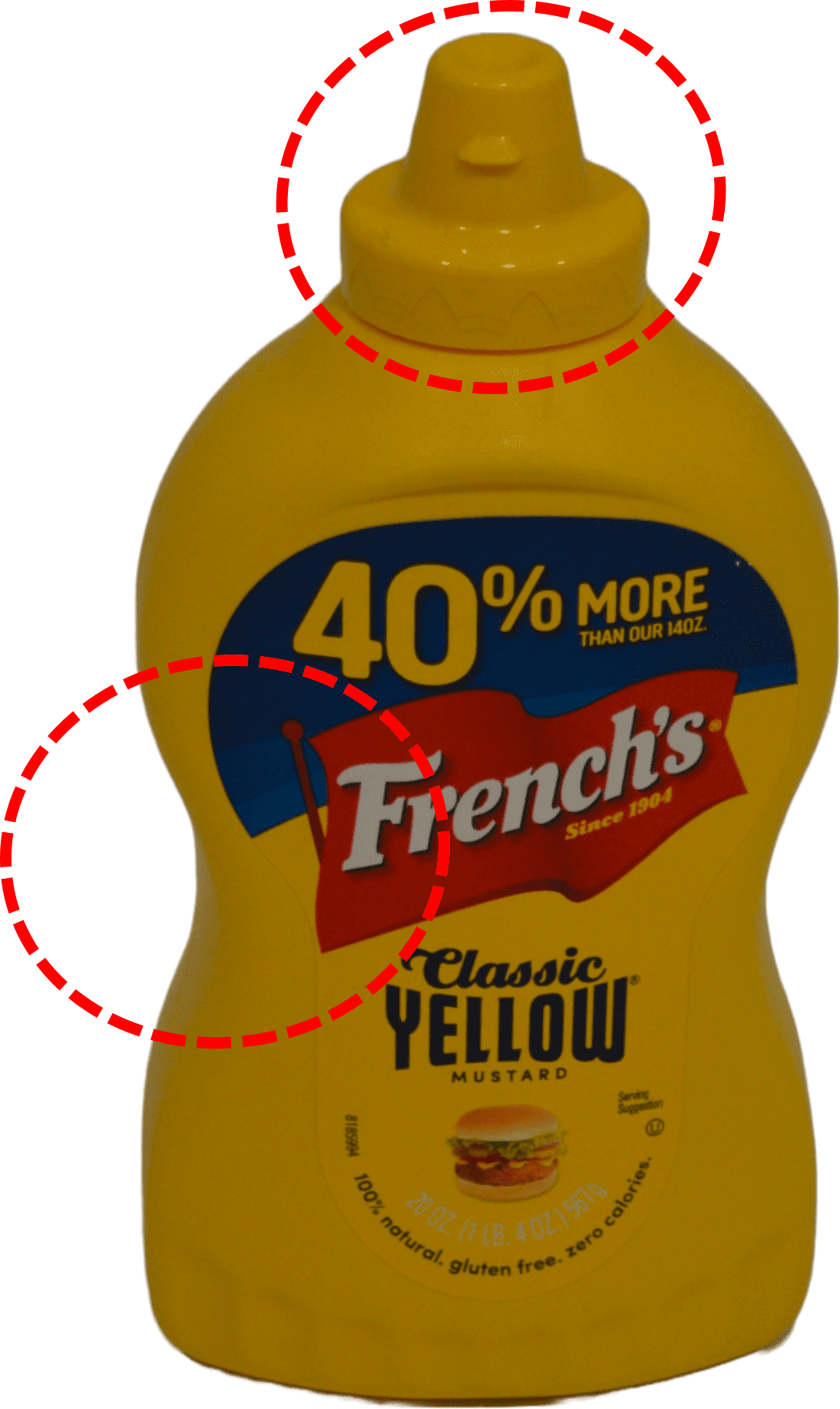}
  \caption{}
  \label{fig:grasp_mustard}
\end{subfigure}%
\begin{subfigure}[t]{.15\textwidth}
  \centering
  \includegraphics[height=0.7\linewidth, keepaspectratio,width=0.7\linewidth]{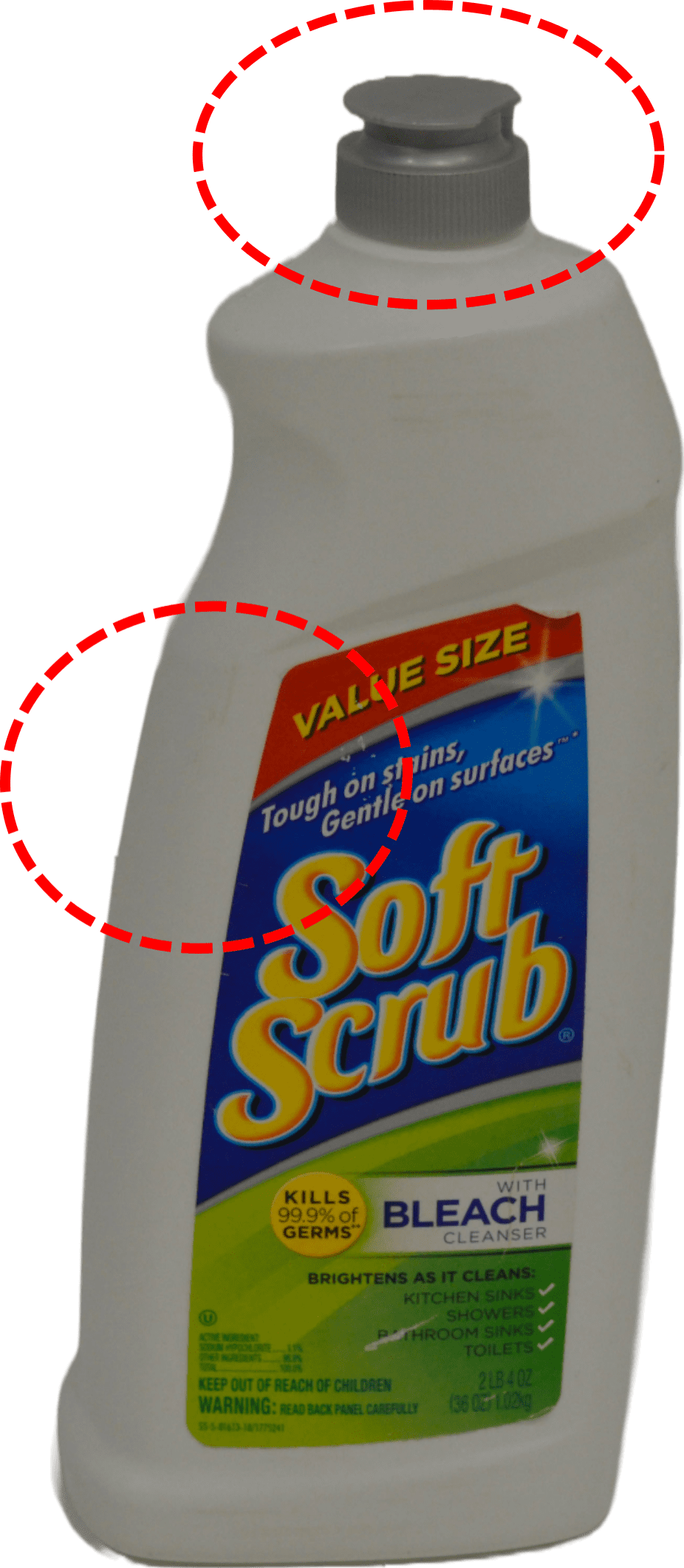}
  \caption{}
  \label{fig:grasp_detergent}
\end{subfigure}%
\begin{subfigure}[t]{.15\textwidth}
  \centering
  \includegraphics[height=0.7\linewidth, keepaspectratio,width=1\linewidth]{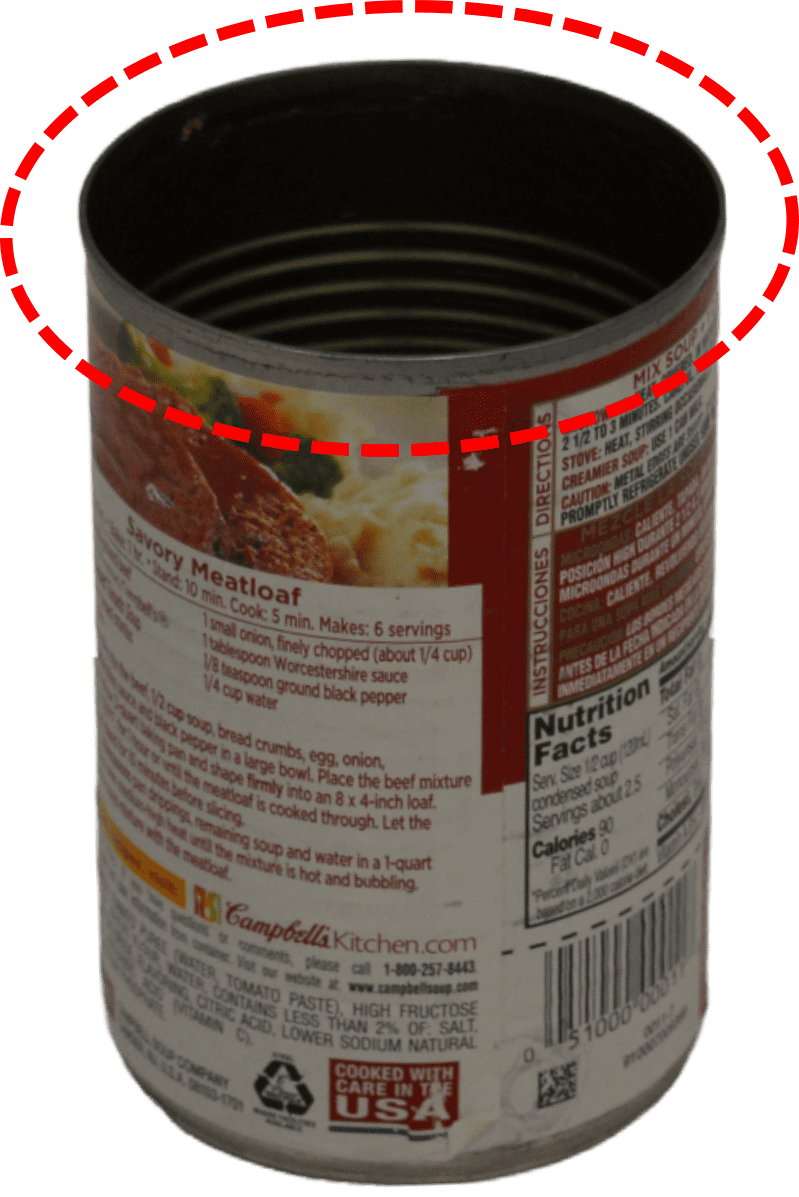}
  \caption{}
  \label{fig:grasp_can}
\end{subfigure}%
\begin{subfigure}[t]{.15\textwidth}
  \centering
  \includegraphics[height=0.7\linewidth, keepaspectratio,width=1\linewidth]{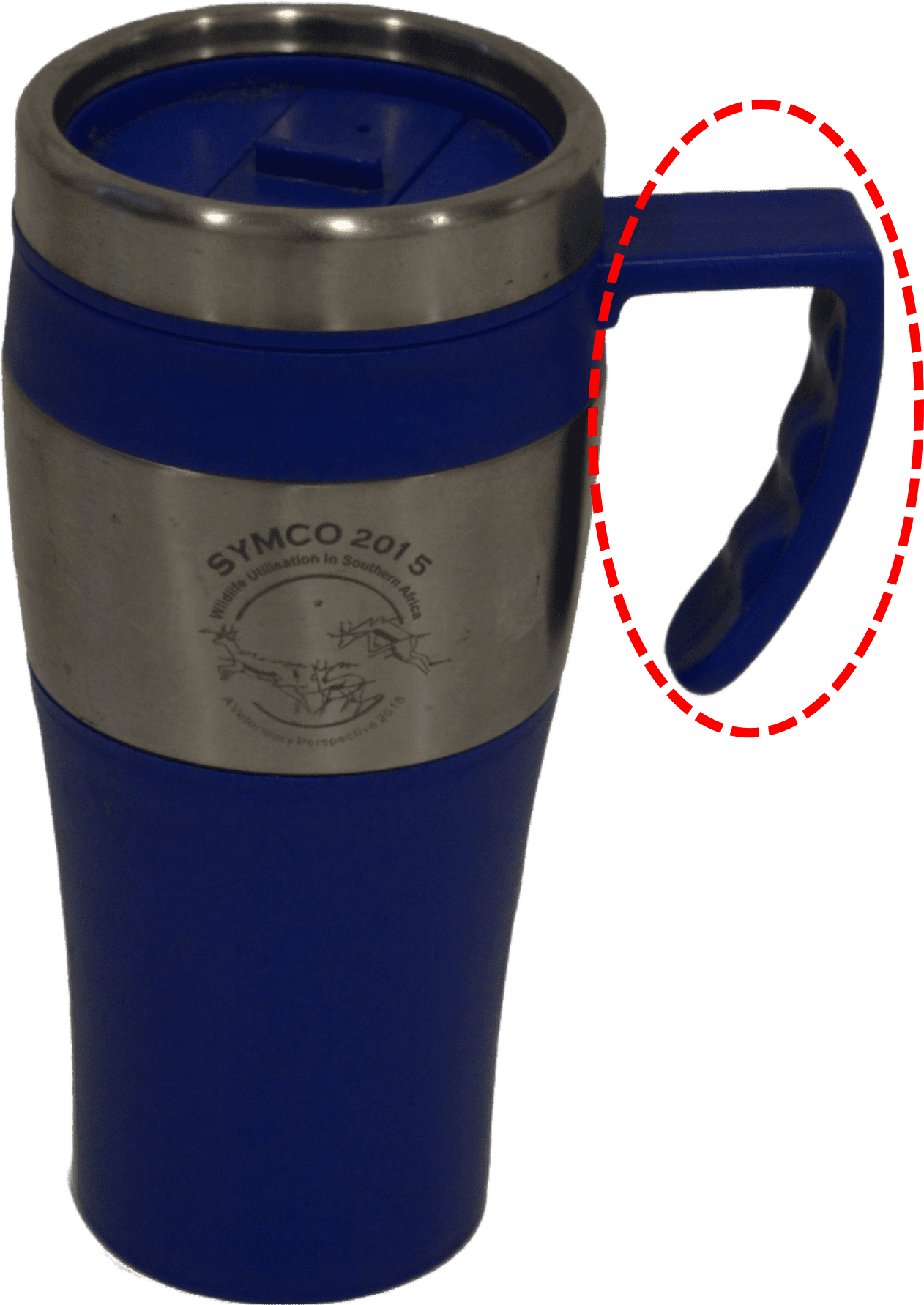}
  \caption{}
  \label{fig:grasp_mug}
\end{subfigure}%
\begin{subfigure}[t]{.15\textwidth}
  \centering
  \includegraphics[height=0.7\linewidth, keepaspectratio,width=1\linewidth]{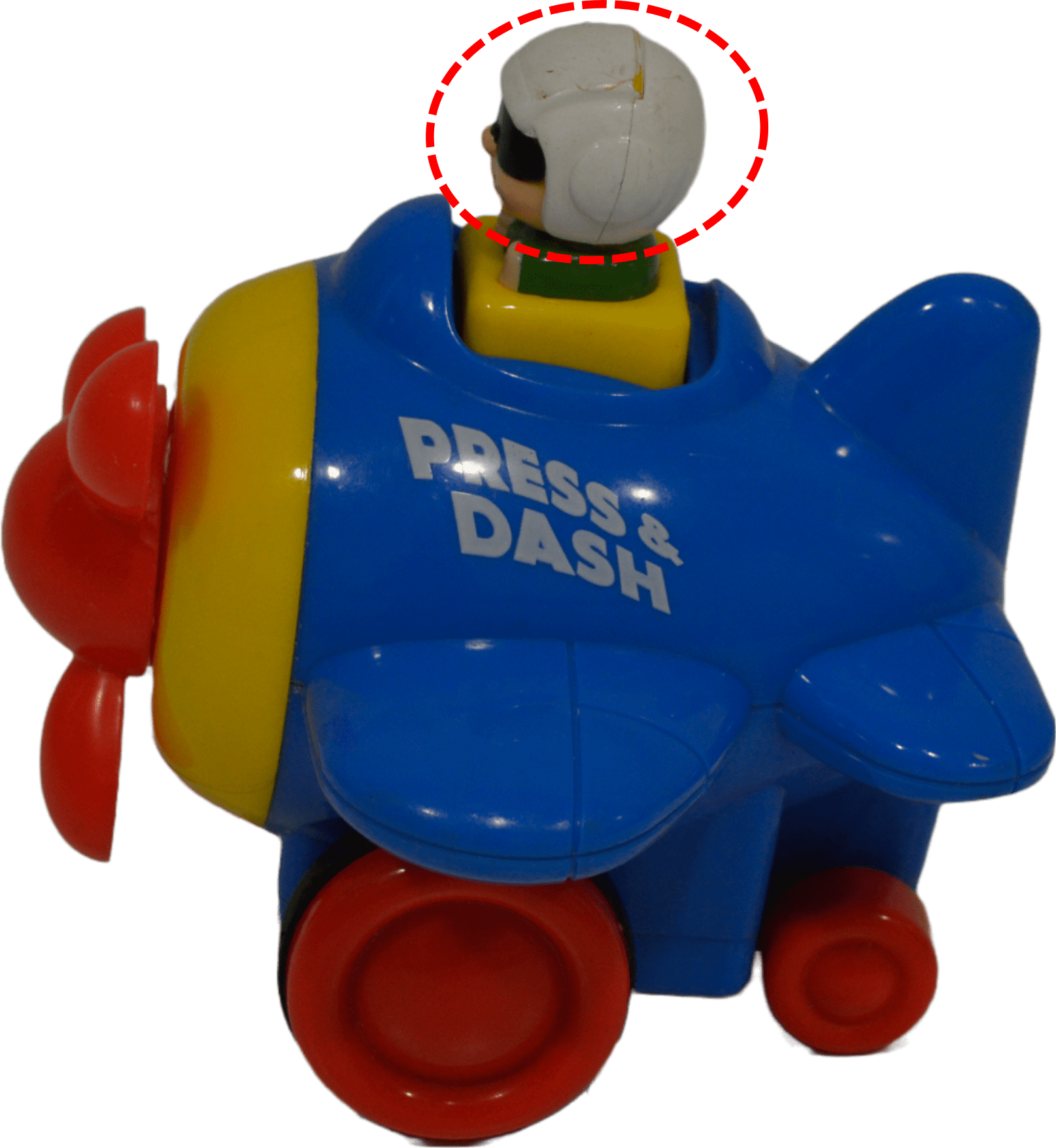}
  \caption{}
  \label{fig:grasp_airplane}
\end{subfigure}%
\begin{subfigure}[t]{.15\textwidth}
  \centering
  \includegraphics[height=0.7\linewidth, keepaspectratio,width=0.5\linewidth]{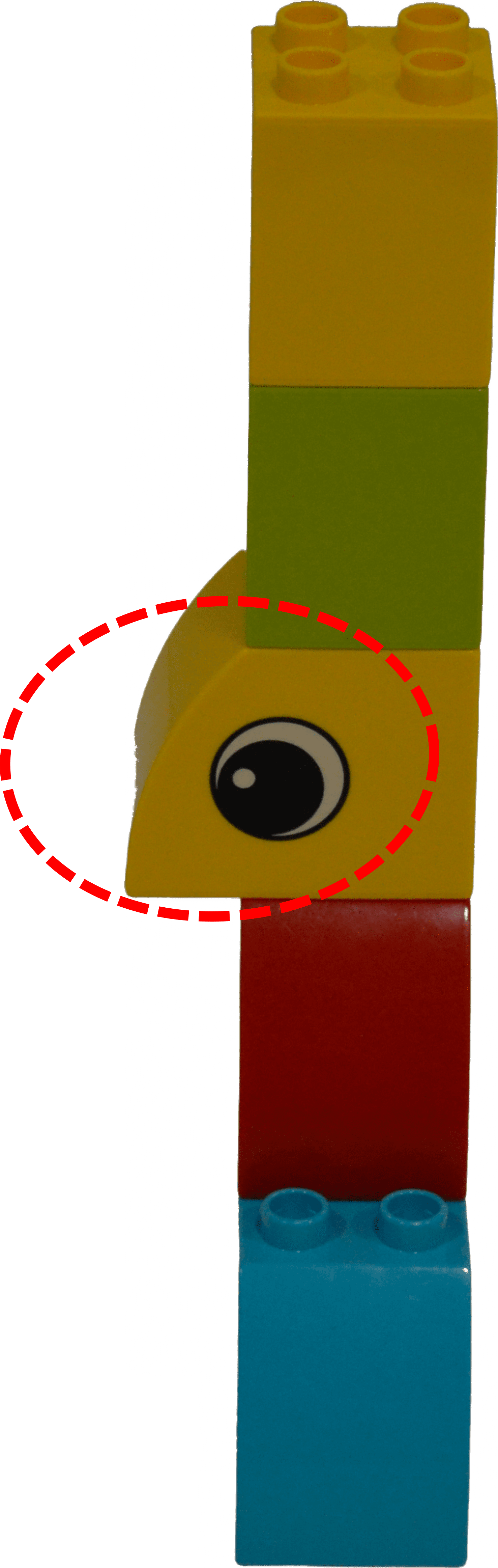}
  \caption{}
  \label{fig:grasp_lego}
\end{subfigure}%
\caption{The 10 objects used in the real-world experiment. All objects, except (a), (j), (k), and (l), are from the YCB object dataset \cite{calliYCBObjectModel2015}. The dashed red lines depict the target grasping area for each object.}
\label{fig:constrained_grasps}
\end{figure*}

In the real robotic experiments, we explored if \ac{methodname} can achieve a higher grasp success rate and be more sample efficient than \graspnet{} when grasping real-world objects. To this end, we used the setup shown in \figref{fig:pull_figure} that included a \franka{} robot to execute the grasps, a \kinect{} to capture the \pcs{}, and an Aruco marker \cite{garrido-juradoAutomaticGenerationDetection2014} for extrinsic calibration.

As the simulation experiments demonstrated that \graspnet{} performs better than \graspnetta{}, we only evaluated \graspnet{} and \ac{methodname}, both of which were trained on synthetic data only, on the 12 different objects and the 16 different target areas presented in \figref{fig:constrained_grasps}. Most of these areas were semantically meaningful and included, for instance, object handles (\figref{fig:grasp_cup}, \figref{fig:grasp_pitcher}, and \figref{fig:grasp_mug}), rims (\figref{fig:grasp_bowl} and \figref{fig:grasp_can}), and caps (\figref{fig:grasp_mustard} and \figref{fig:grasp_detergent}). Each object was placed at a predefined position (see \figref{fig:pull_figure}) but in two different orientations toward the camera: at $0$\textdegree{} and $90$\textdegree. In total, this setup amounted to 32 grasp trials per method. A grasp was successful if the robot picked up the object and moved it to the start pose without dropping it.

To further explore the effect sample size has on grasp success rates, we only sampled 10 grasps at a time per target area. For each batch of 10 grasps, only grasps for which the center grasp point was, at most, 2 cm away from any point in the target area were kept. Each kept grasp was then scored using the evaluation network, and the highest-scoring reachable grasp was executed. If all grasps were removed or no reachable ones were found, another 10 grasps were sampled. The resampling process was repeated until a maximum of 50 grasps had been sampled. If none of the 50 grasps was executed, we terminated the trial and considered the grasp as unsuccessful. Based on this process, the theoretically least number of grasps to be sampled for all objects and target areas together was 320.

The results are presented in \tabref{tb:results_real_world}, and an example grasp is shown in \figref{fig:pull_figure}. These results demonstrate that \ac{methodname} achieves a 15\% higher grasp success rate and is 1.8 times more sample efficiency than \graspnet{}. Both results are also statistically significant. Furthermore, \ac{methodname} only had to sample 40 more grasps than the theoretical lower bound of 320 grasps.

\begin{table}[h!]
    \centering
	\ra{1.3}\tbs{5}
	\caption{\label{tb:results_real_world}The experimental results along with the test statistics and p-values of a pair-wise one-sided Wilcoxon signed-rank test.  $\uparrow$: higher the better, $\downarrow$: lower the better.}
\begin{tabular}{lccl}
	    \toprule
                         & \graspnet{} & \ac{methodname} & \ac{methodname} vs \graspnet{}\\
                                 \midrule
Grasp success rate (\%)~$\uparrow$ & 59.4          & 75.0  &  T=10.0,~~p$<.05^{*}$             \\
\# of grasps sampled~$\downarrow$        & 660            & 360 & T=148.5,~p$<.001^{**}$            \\
        \bottomrule
\end{tabular}
\end{table}


\section{Limitations}
\label{sec:limitations}

In the experimental evaluation, we identified two limitations. The first limitation is that only position constraints can be enforced, while others, such as orientation or reachability constraints, cannot. Although the network did learn implicitly from data which orientations are meaningful to grasp an object successfully, in some scenarios, it would be helpful also explicitly to constrain the orientation of the grasps. As an example, imagine grasping a USB cable for insertion. For this type of object and task, it would make sense to constrain the grasp orientation to not obscure the part of the USB that will be inserted. It would also be helpful to incorporate the robot's reachability as another constraint, as this could further increase sample efficiency by only suggesting reachable areas to grasp.          

The second limitation is refining sampled grasps using the grasp evaluator as was explored in \cite{mousavian6DOFGraspNetVariational2019c}. In that work, sampled grasps were locally refined by moving them in a direction that improved the success probability measured by the grasp evaluator. Unfortunately, as noticed during the experimental evaluation of \cite{mousavian6DOFGraspNetVariational2019c}, such a refining process often moves grasps outside the target area. A possible solution to mitigate this issue would also be to condition the grasp evaluator on the target area\ie{} $\prob(\text{S}\mid\matr{G}, \matr{O}, \matr{A})$, or to constrain the magnitude of the refinement not to leave the target area.

\section{Conclusion}
\label{sec:conclusion}

We presented \ac{methodname}, a generative method for constrained 6-\ac{dof} grasps sampling, and \datasetname{}, a new dataset for learning and evaluating constrained grasp samplers. Constrained grasping has so far been restricted to finding task- or affordance-specific grasps. \ac{methodname} is instead structured to find grasps on any target area whether the area carries task or affordance semantics. The key idea to achieve such general constrained grasping capabilities was to embed the target area as input features for the network and train the network on a dataset that included random target areas of varying sizes. We compared \ac{methodname} to \graspnet{}, a \sota{} generative unconstrained 6-\ac{dof} grasp sampler, in simulation and the real world. The results demonstrate that \ac{methodname} achieves a 10--15\% higher grasp success rate and is 2--3 times more sample efficient than \graspnet{}. 

All in all, the work presented here extends constraint handling in modern neural grasping to arbitrary contact location constraints compared to existing works that constrain grasps primarily from the task compatibility viewpoint. This extension is beneficial as most real-life manipulation tasks pose some constraints on the grasps sampled. However, most tasks require a complex set of constraints to be satisfied. Thus, an important avenue for further research is inferring the constraints from a complex task.

\section*{Acknowledgment}
We gratefully acknowledge the support of NVIDIA Corporation with the donation of the Titan Xp GPU used for this research.

\bibliographystyle{IEEEtran}
\bibliography{refs}  

\begin{thebibliography}{10}
\providecommand{\url}[1]{#1}
\csname url@samestyle\endcsname
\providecommand{\newblock}{\relax}
\providecommand{\bibinfo}[2]{#2}
\providecommand{\BIBentrySTDinterwordspacing}{\spaceskip=0pt\relax}
\providecommand{\BIBentryALTinterwordstretchfactor}{4}
\providecommand{\BIBentryALTinterwordspacing}{\spaceskip=\fontdimen2\font plus
\BIBentryALTinterwordstretchfactor\fontdimen3\font minus
  \fontdimen4\font\relax}
\providecommand{\BIBforeignlanguage}[2]{{%
\expandafter\ifx\csname l@#1\endcsname\relax
\typeout{** WARNING: IEEEtran.bst: No hyphenation pattern has been}%
\typeout{** loaded for the language `#1'. Using the pattern for}%
\typeout{** the default language instead.}%
\else
\language=\csname l@#1\endcsname
\fi
#2}}
\providecommand{\BIBdecl}{\relax}
\BIBdecl

\bibitem{mahlerDexNetDeepLearning2017a}
J.~Mahler, J.~Liang, S.~Niyaz, M.~Laskey, R.~Doan, X.~Liu, J.~Aparicio, and
  K.~Goldberg, ``Dex-{{Net}} 2.0: {{Deep Learning}} to {{Plan Robust Grasps}}
  with {{Synthetic Point Clouds}} and {{Analytic Grasp Metrics}},'' in
  \emph{Robotics: {{Science}} and {{Systems XIII}}}.\hskip 1em plus 0.5em minus
  0.4em\relax {Robotics: Science and Systems Foundation}, Jul. 2017.

\bibitem{morrisonClosingLoopRobotic2018b}
D.~Morrison, J.~Leitner, and P.~Corke, ``Closing the {{Loop}} for {{Robotic
  Grasping}}: {{A Real-time}}, {{Generative Grasp Synthesis Approach}},'' in
  \emph{Robotics: {{Science}} and {{Systems XIV}}}.\hskip 1em plus 0.5em minus
  0.4em\relax {Robotics: Science and Systems Foundation}, Jun. 2018.

\bibitem{mousavian6DOFGraspNetVariational2019c}
A.~Mousavian, C.~Eppner, and D.~Fox, ``6-{{DOF GraspNet}}: {{Variational Grasp
  Generation}} for {{Object Manipulation}},'' in \emph{2019 {{IEEE}}/{{CVF
  International Conference}} on {{Computer Vision}} ({{ICCV}})}.\hskip 1em plus
  0.5em minus 0.4em\relax {Seoul, Korea (South)}: {IEEE}, Oct. 2019, pp.
  2901--2910.

\bibitem{sundermeyerContactGraspNetEfficient6DoF2021b}
M.~Sundermeyer, A.~Mousavian, R.~Triebel, and D.~Fox, ``Contact-{{GraspNet}}:
  {{Efficient}} 6-{{DoF Grasp Generation}} in {{Cluttered Scenes}},'' in
  \emph{2021 {{IEEE International Conference}} on {{Robotics}} and
  {{Automation}} ({{ICRA}})}, May 2021, pp. 13\,438--13\,444.

\bibitem{kokicLearningTaskOrientedGrasping2020a}
M.~Kokic, D.~Kragic, and J.~Bohg, ``Learning {{Task-Oriented Grasping From
  Human Activity Datasets}},'' \emph{IEEE Robotics and Automation Letters},
  vol.~5, no.~2, pp. 3352--3359, Apr. 2020.

\bibitem{detryTaskorientedGraspingSemantic2017}
R.~Detry, J.~Papon, and L.~Matthies, ``Task-oriented grasping with semantic and
  geometric scene understanding,'' in \emph{2017 {{IEEE}}/{{RSJ International
  Conference}} on {{Intelligent Robots}} and {{Systems}} ({{IROS}})}, Sep.
  2017, pp. 3266--3273.

\bibitem{muraliSameObjectDifferent2021}
A.~Murali, W.~Liu, K.~Marino, S.~Chernova, and A.~Gupta, ``Same {{Object}},
  {{Different Grasps}}: {{Data}} and {{Semantic Knowledge}} for {{Task-Oriented
  Grasping}},'' in \emph{Proceedings of the 2020 {{Conference}} on {{Robot
  Learning}}}.\hskip 1em plus 0.5em minus 0.4em\relax {PMLR}, Oct. 2021, pp.
  1540--1557.

\bibitem{liu2020cage}
W.~Liu, A.~Daruna, and S.~Chernova, ``Cage: Context-aware grasping engine,'' in
  \emph{2020 IEEE International Conference on Robotics and Automation
  (ICRA)}.\hskip 1em plus 0.5em minus 0.4em\relax IEEE, 2020, pp. 2550--2556.

\bibitem{brahmbhattContactDBAnalyzingPredicting2019}
S.~Brahmbhatt, C.~Ham, C.~C. Kemp, and J.~Hays, ``{{ContactDB}}: {{Analyzing}}
  and {{Predicting Grasp Contact}} via {{Thermal Imaging}},'' in
  \emph{Proceedings of the {{IEEE}}/{{CVF Conference}} on {{Computer Vision}}
  and {{Pattern Recognition}}}, 2019, pp. 8709--8719.

\bibitem{fangLearningTaskorientedGrasping2020a}
K.~Fang, Y.~Zhu, A.~Garg, A.~Kurenkov, V.~Mehta, L.~{Fei-Fei}, and S.~Savarese,
  ``Learning task-oriented grasping for tool manipulation from simulated
  self-supervision,'' \emph{The International Journal of Robotics Research},
  vol.~39, no. 2-3, pp. 202--216, Mar. 2020.

\bibitem{borstGraspPlanningHow2004}
C.~Borst, M.~Fischer, and G.~Hirzinger, ``Grasp planning: How to choose a
  suitable task wrench space,'' in \emph{{{IEEE International Conference}} on
  {{Robotics}} and {{Automation}}, 2004. {{Proceedings}}. {{ICRA}} '04. 2004},
  vol.~1, Apr. 2004, pp. 319--325 Vol.1.

\bibitem{haschkeTaskorientedQualityMeasures2005}
R.~Haschke, J.~Steil, I.~Steuwer, and H.~Ritter, ``Task-oriented quality
  measures for dextrous grasping,'' in \emph{2005 {{International Symposium}}
  on {{Computational Intelligence}} in {{Robotics}} and {{Automation}}}, Jun.
  2005, pp. 689--694.

\bibitem{kokicAffordanceDetectionTaskspecific2017a}
M.~Kokic, J.~A. Stork, J.~A. Haustein, and D.~Kragic, ``Affordance detection
  for task-specific grasping using deep learning,'' in \emph{2017 {{IEEE-RAS}}
  17th {{International Conference}} on {{Humanoid Robotics}} ({{Humanoids}})},
  Nov. 2017, pp. 91--98.

\bibitem{songLearningTaskConstraints2010}
D.~Song, K.~Huebner, V.~Kyrki, and D.~Kragic, ``Learning task constraints for
  robot grasping using graphical models,'' in \emph{2010 {{IEEE}}/{{RSJ
  International Conference}} on {{Intelligent Robots}} and {{Systems}}}, Oct.
  2010, pp. 1579--1585.

\bibitem{songTaskBasedRobotGrasp2015}
D.~Song, C.~H. Ek, K.~Huebner, and D.~Kragic, ``Task-{{Based Robot Grasp
  Planning Using Probabilistic Inference}},'' \emph{IEEE Transactions on
  Robotics}, vol.~31, no.~3, pp. 546--561, Jun. 2015.

\bibitem{antonovaGlobalSearchBernoulli2018a}
R.~Antonova, M.~Kokic, J.~A. Stork, and D.~Kragic, ``Global {{Search}} with
  {{Bernoulli Alternation Kernel}} for {{Task-oriented Grasping Informed}} by
  {{Simulation}},'' in \emph{Proceedings of {{The}} 2nd {{Conference}} on
  {{Robot Learning}}}.\hskip 1em plus 0.5em minus 0.4em\relax {PMLR}, Oct.
  2018, pp. 641--650.

\bibitem{fangGraspNet1BillionLargeScaleBenchmark2020}
H.-S. Fang, C.~Wang, M.~Gou, and C.~Lu, ``{{GraspNet-1Billion}}: {{A
  Large-Scale Benchmark}} for {{General Object Grasping}},'' in \emph{2020
  {{IEEE}}/{{CVF Conference}} on {{Computer Vision}} and {{Pattern
  Recognition}} ({{CVPR}})}.\hskip 1em plus 0.5em minus 0.4em\relax {Seattle,
  WA, USA}: {IEEE}, Jun. 2020, pp. 11\,441--11\,450.

\bibitem{eppnerACRONYMLargeScaleGrasp2021a}
C.~Eppner, A.~Mousavian, and D.~Fox, ``{{ACRONYM}}: {{A Large-Scale Grasp
  Dataset Based}} on {{Simulation}},'' in \emph{2021 {{IEEE International
  Conference}} on {{Robotics}} and {{Automation}} ({{ICRA}})}, May 2021, pp.
  6222--6227.

\bibitem{lundellMultiFinGANGenerativeCoarseToFine2021a}
J.~Lundell, E.~Corona, T.~Nguyen~Le, F.~Verdoja, P.~Weinzaepfel, G.~Rogez,
  F.~{Moreno-Noguer}, and V.~Kyrki, ``Multi-{{FinGAN}}: {{Generative
  Coarse-To-Fine Sampling}} of {{Multi-Finger Grasps}},'' in \emph{2021 {{IEEE
  International Conference}} on {{Robotics}} and {{Automation}} ({{ICRA}})},
  May 2021, pp. 4495--4501.

\bibitem{depierreJacquardLargeScale2018b}
A.~Depierre, E.~Dellandr{\'e}a, and L.~Chen, ``Jacquard: {{A Large Scale
  Dataset}} for {{Robotic Grasp Detection}},'' in \emph{2018 {{IEEE}}/{{RSJ
  International Conference}} on {{Intelligent Robots}} and {{Systems}}
  ({{IROS}})}, Oct. 2018, pp. 3511--3516.

\bibitem{leDeformationAwareDataDrivenGrasp2022}
T.~N. Le, J.~Lundell, F.~J. {Abu-Dakka}, and V.~Kyrki, ``Deformation-{{Aware
  Data-Driven Grasp Synthesis}},'' \emph{IEEE Robotics and Automation Letters},
  vol.~7, no.~2, pp. 3038--3045, Apr. 2022.

\bibitem{EppnerISRR2019}
C.~Eppner, A.~Mousavian, and D.~Fox, ``A billion ways to grasps - an evaluation
  of grasp sampling schemes on a dense, physics-based grasp data set,'' in
  \emph{Proceedings of the International Symposium on Robotics Research
  ({ISRR})}, Hanoi, Vietnam, 2019.

\bibitem{morrisonEGADEvolvedGrasping2020}
D.~Morrison, P.~Corke, and J.~Leitner, ``{{EGAD}}! {{An Evolved Grasping
  Analysis Dataset}} for {{Diversity}} and {{Reproducibility}} in {{Robotic
  Manipulation}},'' \emph{IEEE Robotics and Automation Letters}, vol.~5, no.~3,
  pp. 4368--4375, Jul. 2020.

\bibitem{lenzDeepLearningDetecting2015a}
I.~Lenz, H.~Lee, and A.~Saxena, ``Deep learning for detecting robotic grasps,''
  \emph{The International Journal of Robotics Research}, vol.~34, no. 4-5, pp.
  705--724, Apr. 2015.

\bibitem{sohnLearningStructuredOutput2015}
K.~Sohn, H.~Lee, and X.~Yan, ``Learning {{Structured Output Representation}}
  using {{Deep Conditional Generative Models}},'' in \emph{Advances in {{Neural
  Information Processing Systems}}}, vol.~28.\hskip 1em plus 0.5em minus
  0.4em\relax {Curran Associates, Inc.}, 2015.

\bibitem{qiPointNetDeepHierarchical2017a}
C.~R. Qi, L.~Yi, H.~Su, and L.~J. Guibas, ``{{PointNet}}++: {{Deep Hierarchical
  Feature Learning}} on {{Point Sets}} in a {{Metric Space}},'' in
  \emph{Advances in {{Neural Information Processing Systems}}}, vol.~30.\hskip
  1em plus 0.5em minus 0.4em\relax {Curran Associates, Inc.}, 2017.

\bibitem{makoviychukIsaacGymHigh2021a}
V.~Makoviychuk, L.~Wawrzyniak, Y.~Guo, M.~Lu, K.~Storey, M.~Macklin,
  D.~Hoeller, N.~Rudin, A.~Allshire, A.~Handa, and G.~State, ``Isaac {{Gym}}:
  {{High Performance GPU-Based Physics Simulation For Robot Learning}},'' Aug.
  2021.

\bibitem{calliYCBObjectModel2015}
B.~Calli, A.~Singh, A.~Walsman, S.~Srinivasa, P.~Abbeel, and A.~M. Dollar,
  ``The {{YCB}} object and {{Model}} set: {{Towards}} common benchmarks for
  manipulation research,'' in \emph{2015 {{International Conference}} on
  {{Advanced Robotics}} ({{ICAR}})}, Jul. 2015, pp. 510--517.

\bibitem{garrido-juradoAutomaticGenerationDetection2014}
S.~{Garrido-Jurado}, R.~{Mu{\~n}oz-Salinas}, F.~J. {Madrid-Cuevas}, and M.~J.
  {Mar{\'i}n-Jim{\'e}nez}, ``Automatic generation and detection of highly
  reliable fiducial markers under occlusion,'' \emph{Pattern Recognition},
  vol.~47, no.~6, pp. 2280--2292, Jun. 2014.

\end{thebibliography}

\end{document}